\documentclass[10pt,twocolumn,letterpaper]{article}

\usepackage{cvpr}
\usepackage{times}
\usepackage{epsfig}
\usepackage{graphicx}
\usepackage{amsmath}
\usepackage{amssymb}
\usepackage{subfig}
\usepackage{authblk}
\usepackage[breaklinks=true,bookmarks=false]{hyperref}
\cvprfinalcopy % *** Uncomment this line for the final submission

\def\cvprPaperID{2348} % *** Enter the CVPR Paper ID here

\ifcvprfinal\fi
\title{Simulations for Validation of Vision Systems}
\author[1]{V S R Veeravasarapu\thanks{subbu@fias.uni-frankfurt.de}}
\author[1]{Rudra Narayan Hota\thanks{hota@fias.uni-frankfurt.de}}
\author[2]{Constantin Rothkopf \thanks{rothkopf@psychologie.tu-darmstadt.de}}
\author[1]{Ramesh Visvanathan \thanks{ramesh@fias.uni-frankfurt.de}}
\affil[1]{Center for Cognition and Computation, Goethe University Frankfurt}
\affil[2]{Institute of Psychology, Technical University Darmstadt}

\begin{document}
\maketitle
%\thispagestyle{empty}

%%%%%%%%% ABSTRACT
\begin{abstract}
As the computer vision matures into a systems science and engineering discipline, there is a trend in leveraging latest advances in computer graphics simulations for performance evaluation, learning, and inference. However, there is an open question on the utility of graphics simulations for vision with apparently contradicting views in the literature. In this paper, we place the results from the recent literature in the context of performance characterization methodology outlined in the 90's and note that insights derived from simulations can be qualitative or quantitative depending on the degree of fidelity of models used in simulation and the nature of the question posed by the experimenter. 
We describe a simulation platform that incorporates latest graphics advances and use it for systematic performance characterization and tradeoff analysis for vision system design. We verify the utility of the platform in a case study of validating a generative model inspired vision hypothesis, Rank-Order consistency model, in the contexts of global and local illumination changes, and bad weather, and high-frequency noise. Our approach establishes the link between alternative viewpoints, involving models with physics based semantics
and signal and perturbation semantics and confirms insights in literature on robust change detection.
\end{abstract}

%%%%%%%%% BODY TEXT

\section{Introduction}

\begin{figure*}
\centering
\includegraphics[width=4.25cm, height=3cm]{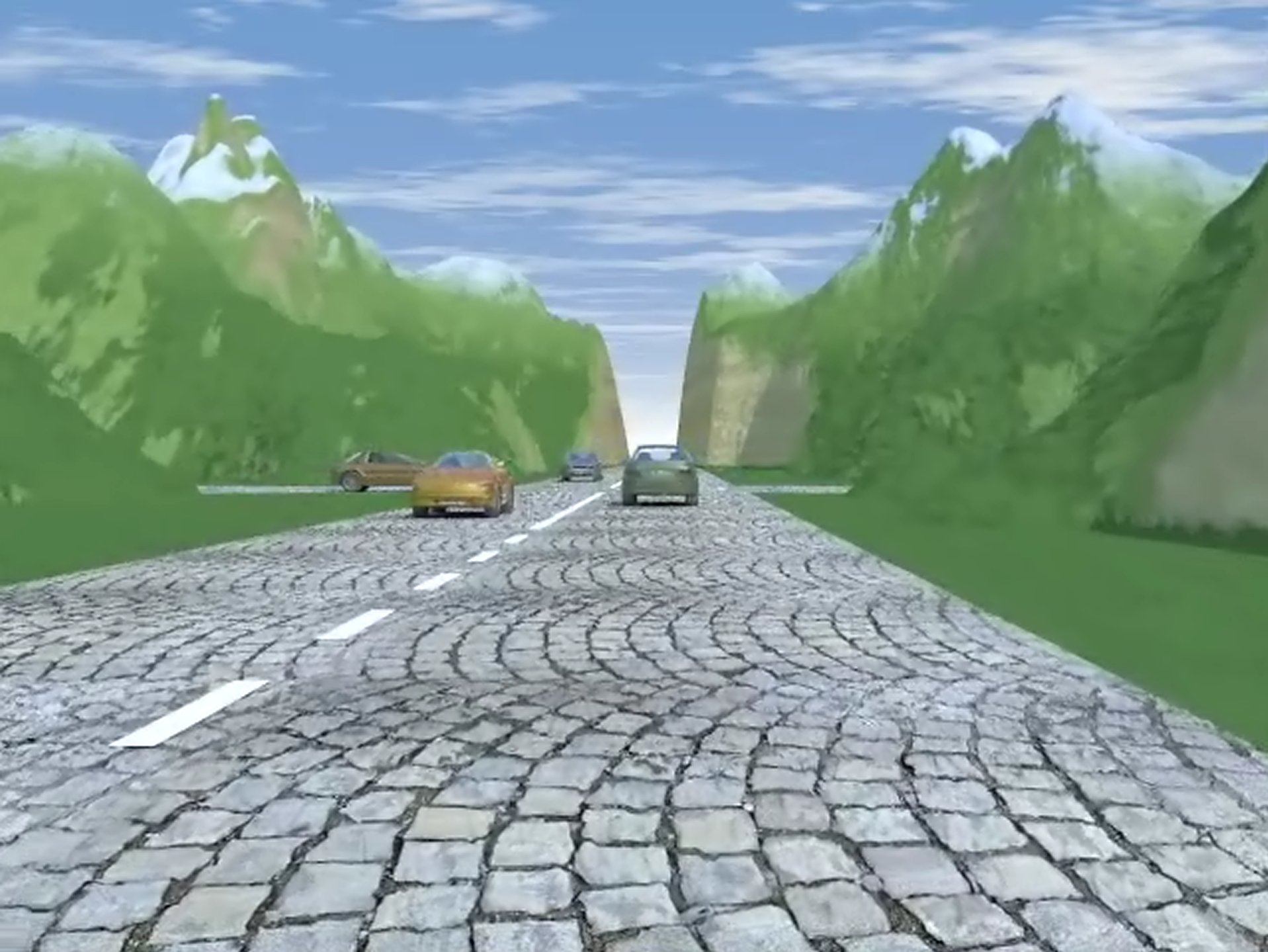}
\includegraphics[width=4.25cm, height=3cm]{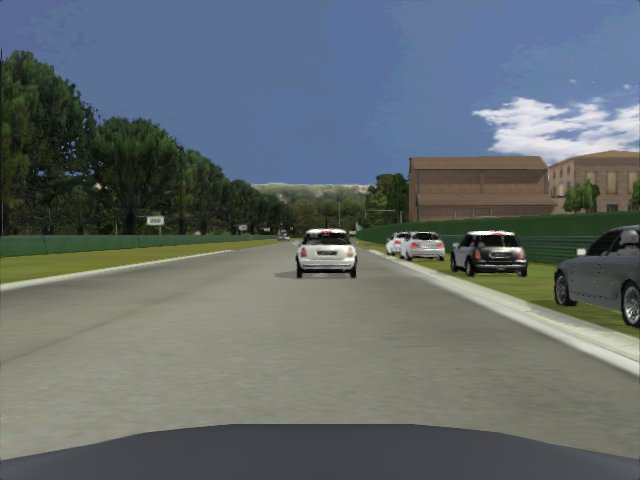}
\includegraphics[width=4.25cm, height=3cm]{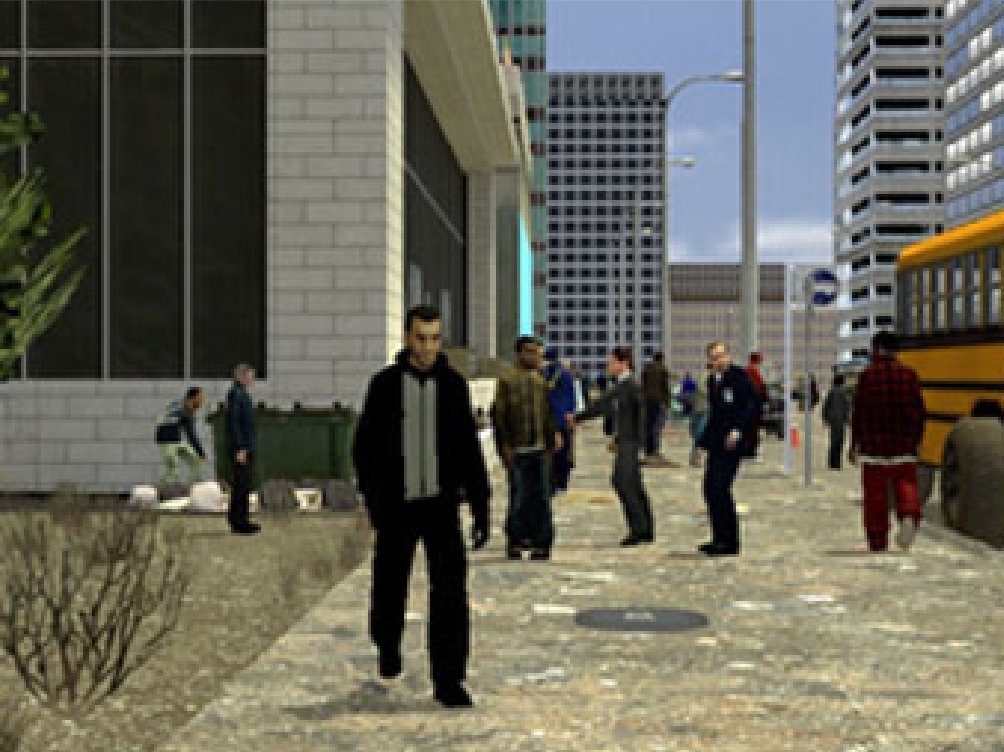}
\includegraphics[width=4.25cm, height=3cm]{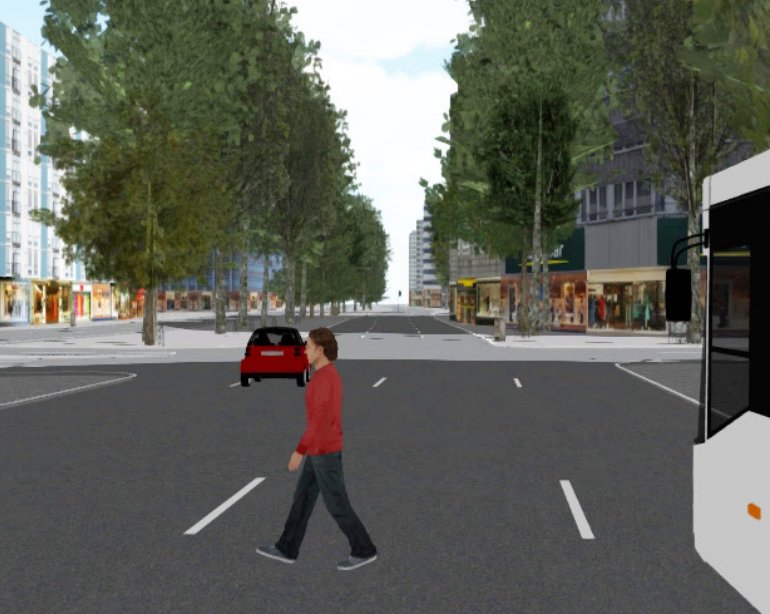}
\includegraphics[width=4.25cm, height=3cm]{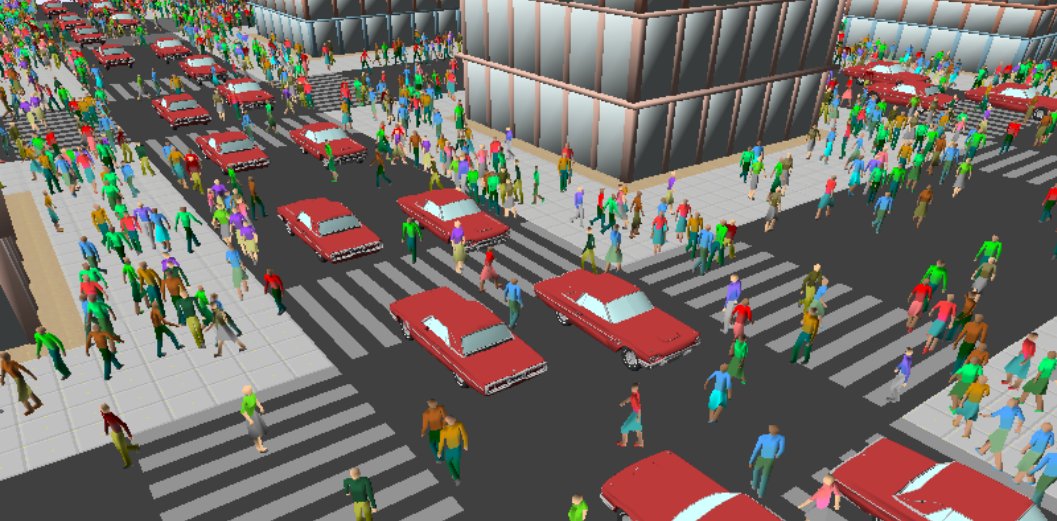}
\includegraphics[width=4.25cm, height=3cm]{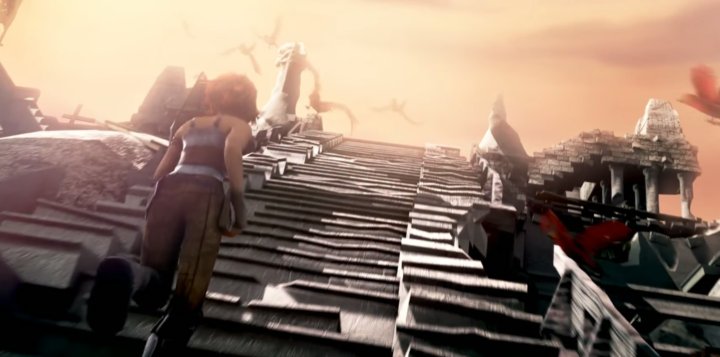}
\includegraphics[width=4.25cm, height=3cm]{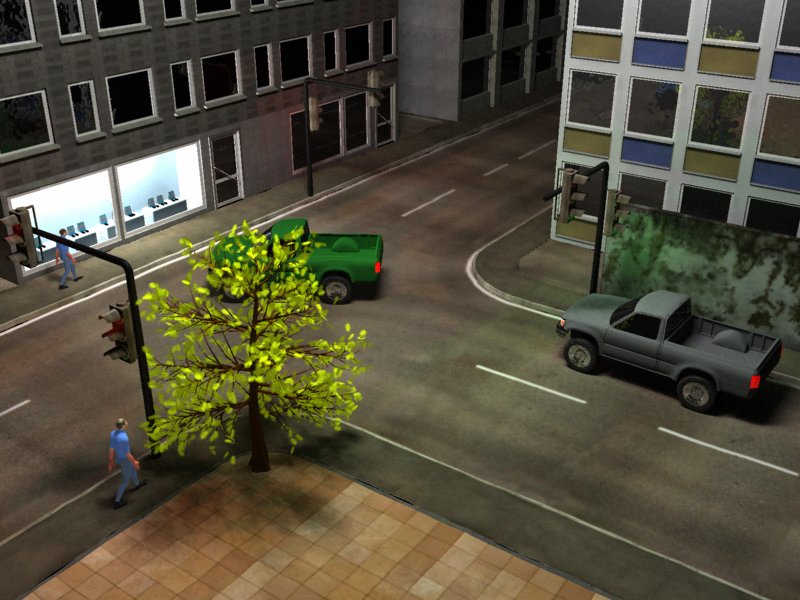}
\includegraphics[width=4.25cm, height=3cm]{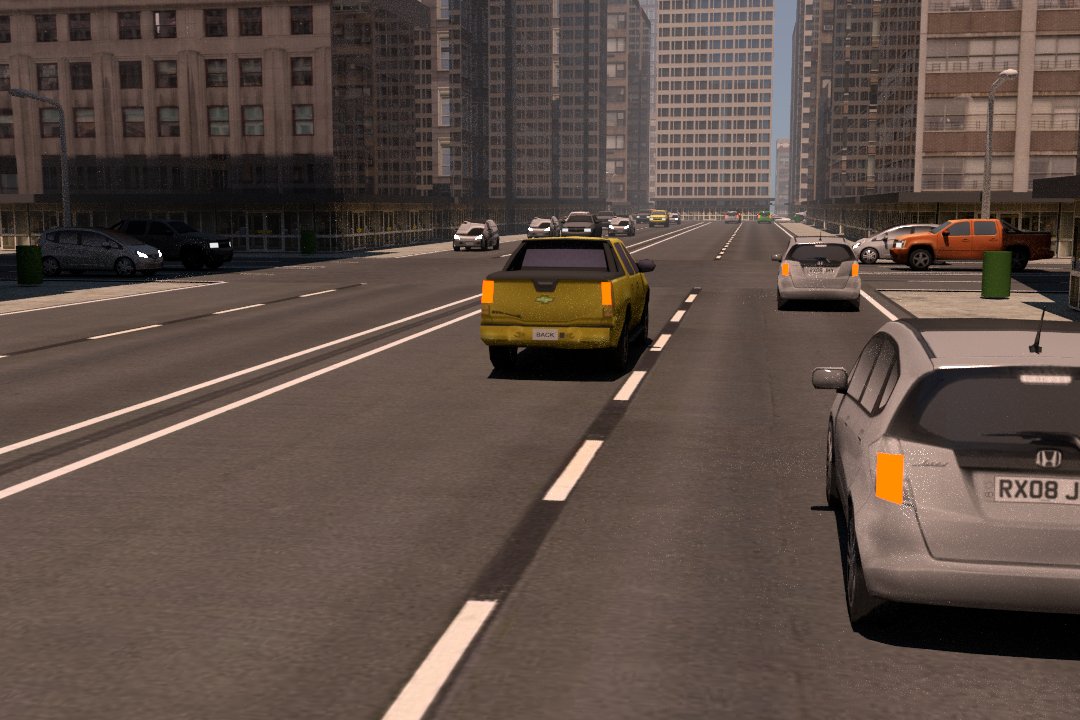}
\caption{\small{Comparision with existing simulators: \textit{Top row (left to right)} Simulated data used in \cite{vaudrey2008differences}, Vdrift Simulator \cite{haltakov2013framework}, Half-Life simulator \cite{vazquez2014virtual,DBLP:journals/pami/XuRVL14}, ADAS simulator; 
\textit{Bottom row (left to right)}:- Crowd simulator, MPI-Sintel data \cite{butler2012naturalistic}, SABS \cite{cvpr11brutzer} atrificial data for background substraction and Our simulator}}
\label{fig_compare_existing_sim}
\end{figure*}

% general discussion on role of graphics simulation and performance characterization viewpoint

Computer simulations can play a dominant role in evaluating the behavior of alternative implementations and in systematic performance evaluation and validation. Recently, model-based simulations are increasingly used in systems engineering and integrated into machine learning or probabilistic programming platforms \cite{mansinghka2013approximate} that allow numerical simulations to be tightly integrated into learning and inference. This opens up fundamental questions such as: What is the value of the synthetic imagery in experimental computer vision? How much fidelity or realism is needed?, etc. In our opinion, these questions haven't yet been answered properly in the literature. From a performance characterization standpoint \cite{bb3334}, a graphics simulation engine's usefulness can be evaluated by thinking of it as a parameterized system whose input involving attributes such as: scene and object geometry, appearance, illumination, dynamics, environment, sensor and rendering parameters etc.,  are translated to image or video output. Deviations from reality in the inputs along with nature of computation used in the simulation engine map to deviations in the rendered output. These deviations in rendered data propagate through the subsequent stage to produce deviations in the final output.  The significance of the impact of these deviations on the experimental conclusion depends on the nature of the conclusions an experimenter wishes to draw. These conclusions may range from qualitative to quantitative aspects of a vision system. The effect of the degree of fidelity on these conclusions would vary depending on (a) the nature of information/features that are being propagated from graphics to vision system, (b) their invariance to other scene and graphics parameters and (c) the closeness of the physics processes and models(that directly influences feature rendering) to reality. For a systematic study of the value of graphics simulations to vision w.r.t. these issues, we focus on developing a flexible parametric generative system based on physics based graphics, along with image annotating shaders. We demonstrate the use of this framework to illustrate varied ways of utilization of simulation for drawing conclusions about vision systems performance.

% skepticism
In the past, vision community has been skeptical about using graphics simulations for the design of vision algorithms/systems because the contemporary graphics was only able to simulate highly approximate and non-realistic renderings. The argument that has been brought up often, was that synthetic images are simulated based on approximations which are trivial for vision algorithms to solve. This might be true for the simulations rendered with local illumination models (see Fig.\ref{fig_compare_existing_sim}.a-f and Fig.\ref{fig_fidelity}.a, which are rendered using Blin-Phong reflection shader). For example, the classical assumptions in the optical flow literature, such as \textit{Brightness constancy} or \textit{linear transformations} to derive data term and \textit{local flow smoothness assumptions} to derive a penalty term or regularizer in energy minimization frameworks, are violated often in real world datasets. There is a diverse set of graphics rendering tools addressing varied degrees of photorealism requirements (e.g. graphics for movies vs gaming etc).

% organization of rest of paper

\textbf{Organization}: Section \ref{sec_graphics_for_vision} consolidates few related works which exploit and/or evaluate the utility of graphics in vision system design process, and summarizes the contributions of this work. Section \ref{sec_graphics_framework} provides a short overview of our simulation framework. In Section \ref{sec_hyp_validate}, we use the framework to validate a vision hypothesis, e.g. Rank-Order consistency model, and use our simulation platform to conduct controlled experiments involving different temporal contexts: global ilumination change, local illumination change, bad weather (varied fog density). The experiments provide  both qualitative and quantitative insights into how algorithms perform on simulated data when compared with data from a real-world change detection database. Section \ref{sec_changedetect} demonstrates how these experiments can be augmented with more classical signal and perturbation simulations based performance characterization. The combined viewpoints provide broader insights on invariance properties and allows for establishment of  
explicit mapping between physics based contextual models and algorithms that have generative model-based semantics in the context of change detection in video. Section \ref{sec_conc} provides some conclusions and future efforts. 
The supplementary material provided describes more details of the graphics simulation platform. In addition, we provide details about change detection algorithms and the application context.

\begin{table*}
\footnotesize
\centering
\begin{tabular}{|l|p{3.4cm}|p{3.4cm}|p{3.4cm}|p{3.4cm}|} \hline
Work & Data-Generation-phase & Vision-task & Inferences & Conclusions $\&$ Remarks \\ \hline

\cite{parameswaran2012design} 2012 & Used 3D CAD models of domain geometry and basic rendering algorithm  & \textbf{\textit{Learning from Simulations}}: To estimate distributions of people count given crowdness and optimal camera setup in the site & \textbf{\textit{Quantitative analysis}}: Used the models in the real world queue statistics estimation system & \textbf{Useful}: Existing 3D CAD models can be used with basic rendering algos to modeling purposes. \\ \hline

\cite{vaudrey2008differences} 2008 & Used very basic shading algorithms to create the data & \textbf{\textit{Learning from Simulations}}: Trained the optical flow and disparity matching systems on the simulated data  & \textbf{\textit{Quantitative analysis}}: Used the trained models on the real world data and found counter-effective results & \textbf{Not useful}: Graphics are too trivial to train the vision algorithms \\ \hline

\cite{Kaneva_2011} 2011 & Used a photo-realistic city model to generate the data  & \textbf{\textit{Learning from Simulations}}: Trained the different feature descriptors on the simulated data & \textit{\textbf{Qualitative analysis}}: Ranked the features  & \textbf{Useful}: The rankings are same as that of real world.\\ \hline

\cite{vazquez2014virtual,DBLP:journals/pami/XuRVL14} 2014 & Used an existing game with near realistic 3D models to generate the labelled data & \textbf{\textit{Learning from Simulations}}: Trained pedestrian detectors based on different feature descriptors HOG, LBP and HOG+LBP & \textbf{\textit{Quantitative analysis}}: Added few real worlds samples in training phase and achieved best performance for LBP and HOG+LBP on real data & \textbf{Useful}: Graphics can be used and bias in the results can be corrected with transfer learning concepts. \\ \hline

\cite{fischer2015flownet} 2015 & Used a combination of real image backgrounds and synthetic object 3D models with random affine movements & \textbf{\textit{Learning from Simulations}}: Trained a CNN for optical flow estimation:  & \textbf{\textit{Quantitative analysis}}: Used the trained network on real data & \textbf{Useful}: Achieved state-of-the-art performance \\ \hline

\cite{mansinghka2013approximate, kulkarnipicture} 2013 & Basic rendering algos to simulate feature likelihoods & \textbf{\textit{Inference process using Simulations}}: Simulated features are accepted or rejected based on similarity to the input image & \textbf{\textit{Qualitative analysis}}: Features and stochastic thresholding step provide invariance to fidelity  & \textbf{Useful}: More realistic models and graphics can bootstrap the approximate inferences \\ \hline

\cite{meister2011real} 2011 & Used carefully designed indoor scene and global rendering algorithm for generation & \textbf{\textit{Performance Characterization by Simulations}}  Used the data to validate optical flow algorithms & \textbf{\textit{Quantitative analysis}}: Error locations and average error metrics are compared against real world data & \textbf{Useful}: Some deviations are observed in the spatial locations of errors while average errors are similar.\\ \hline

%\cite{butler2012naturalistic} 2012 & used open source animated movie for data generation with global rendering algo  & \textbf{\textit{Validation by Simulations}}:  & Quantitative analysis & \\ \hline

\end{tabular}
\caption{\small{Table illustrates the seemingly diverging conclusions in recent work. We note, that all of these observations are valid given their context and degree of model fidelity used for simulations.  From a performance characterization standpoint, the deviations in rendered data propagate through the subsequent stage to produce deviations in the final output. Thus, the significance of the impact of these deviations on the experimental conclusion depends on the nature of the conclusions (i.e. qualitative or quantitative) drawn and the degree of correctness or fidelity of models used in the simulation.}}
\label{tab:unification}
\end{table*}

\section{Literature Review} \label{sec_graphics_for_vision}
Although, whether graphics simulations are in fact realistic enough for computer vision remains an open question, graphics has been used for different purposes in the development cycle of vision algorithms, including performance modeling and improvements, parameter learning, transfer learning and inference etc.
%\textbf{WRITE SOMETHING ABOUT TABLE 1 and support it from a coherent view}
 Table.\ref{tab:unification} consolidates the most relevant works that evaluated the use of graphics simulations for different purposes. 
%Because of the diversity of vision algorithms and models used for simulations, all these works provide diverging conclusions. 
The works \cite{vaudrey2008differences,vazquez2014virtual,DBLP:journals/pami/XuRVL14} used basic rendering algorithms and scene parameters and concluded that simulations are not useful for tuning the respective vision systems, where as the work \cite{meister2011real} used carefully designed indoor scene models (parameters) and advanced rendering algorithms to synthesize very realistic sensory data and concluded that graphics can be used. The work \cite{butler2012naturalistic} showed that motion models and local spatial statistics, crucial for optic flow estimation, match with reality. Hence, they argued that the Sintel data could be used to design and tune the flow estimators even though the data is not photo-realistic. Recently, \cite{fischer2015flownet} used a combination of real-data and synthetic objects to train a CNN and provide empirical evidence of the usefulness of such training for real world settings. The main focus of these works is in demonstration of the utility or lack of utility of simulation for vision systems design and the emphasis is on evaluation of the system as a black-box in an application context. We note that all these observations may be explained from a unified perspective inspired from performance characterization. Performance characterization of an algorithm/system is defined as relating the output deviations to the statistics of ideal input, input deviations and system/algorithm's free parameters. Likewise, the deviations in system's conclusions due to the realism of simulated data used for tuning/training, could be related to closeness of the models used on scene (virtual world) and graphics (rendering pipeline) to reality and also to the choice of vision algorithms and their parameters.

%\textcolor{red}{
%\subsection{Key Contributions} \label{sec_contributions}
In summary, our main contributions in the paper include: 
\begin{itemize}
\item{} We place the results from recent literature in the context of performance characterization methodology developed in the 90's.  
%Performance characterization of an algorithm/system is defined as relating the output deviations to the statistics of ideal input, input deviations and system/algorithm's free parameters. Likewise, the deviations in system's conclusions due to the realism of simulated data used for tuning/training, could be related to closeness of the models used on scene (virtual world) and graphics (rendering pipeline) to reality and also to the choice of vision algorithms and their parameters. In other words, the effect of the degree of fidelity on these conclusions would vary depending on (a) the nature of information/features that are being propagated from graphics to vision system, (b) their invariance to other scene and graphics parameters and (c) the closeness of the physics processes and models(that directly influences feature rendering) to reality.  
The combination of model-based characterization along with the latest computer graphics simulations is the novelty in our work. 
\item{} The development of a platform integrating physics based models and state-of-the-art rendering algorithms, motivated by the need to analyze the effects of photo-realism or feature realism of the data on system characterization.
% We motivated the need of analyzing the effects of photo-realism or feature realism of the data on system characterization conclusion. Thus, we need a platform that is integrated with physics based models and state-of-the-art rendering algorithms. The development of the platform is one of the contributions. 
\item{} Finally, simulation based systems characterization gives varied degrees of insights from establishing correctness of implementation, to providing qualitative and quantitative insights. We demonstrate the utility of the platform to provide qualitative to quantitative assessments of performance in a specific case study involving change detection. %The case study validates an hypothesis from generative model based vision literature  while another case study highlights how the evaluation of algorithms that have generative model-based semantics but lacking correspondence to physics-based models can be achieved by use of the simulation platform.
\end{itemize}
%}

\section{Graphics Simulation Platform }\label{sec_graphics_framework}

\begin{figure*}
\includegraphics[width=5.7cm,height=3.8cm]{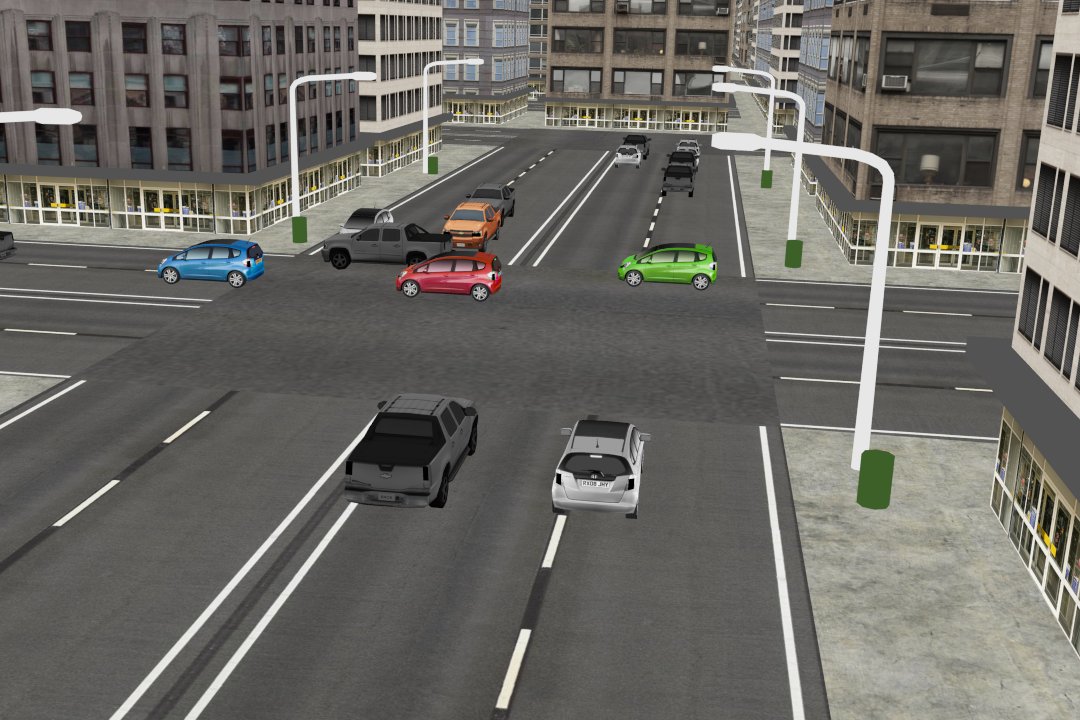}
\includegraphics[width=5.7cm,height=3.8cm]{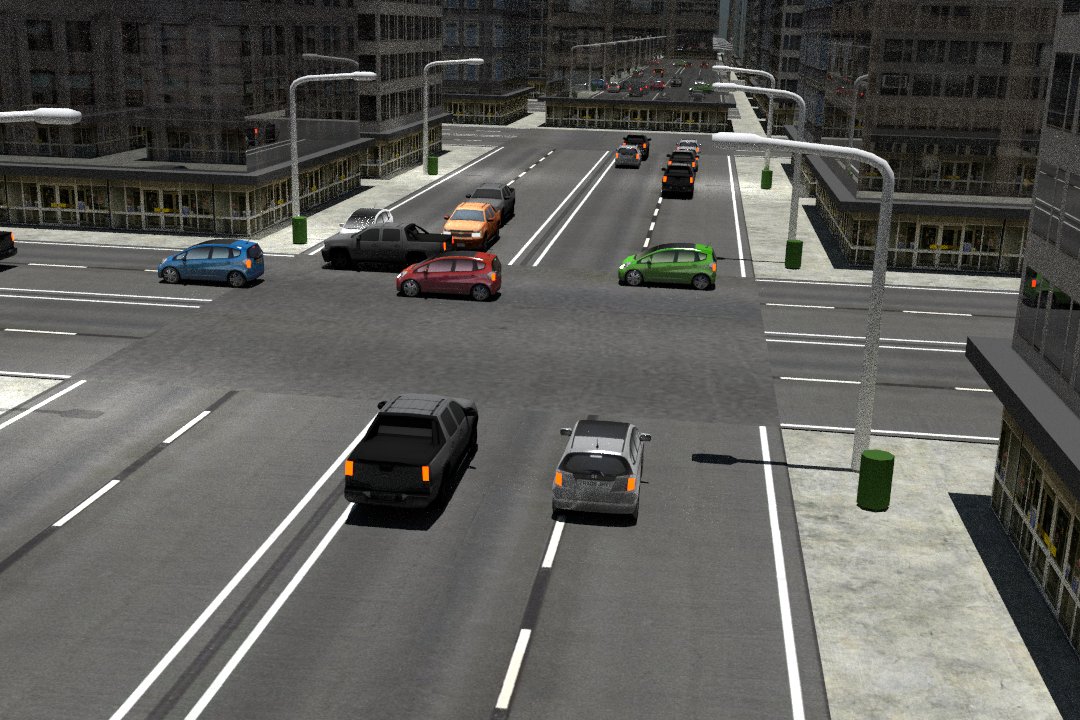}
\includegraphics[width=5.7cm,height=3.8cm]{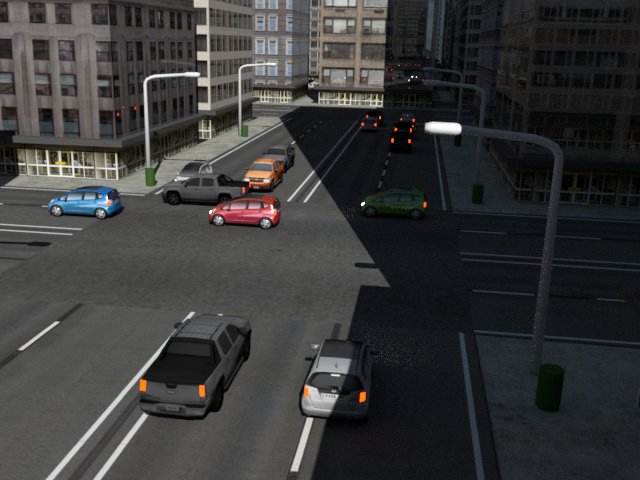} \\
\includegraphics[width=5.7cm,height=3.8cm]{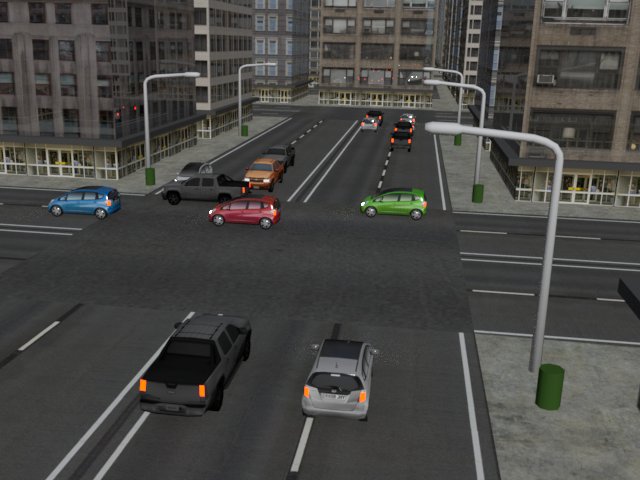}
\includegraphics[width=5.7cm,height=3.8cm]{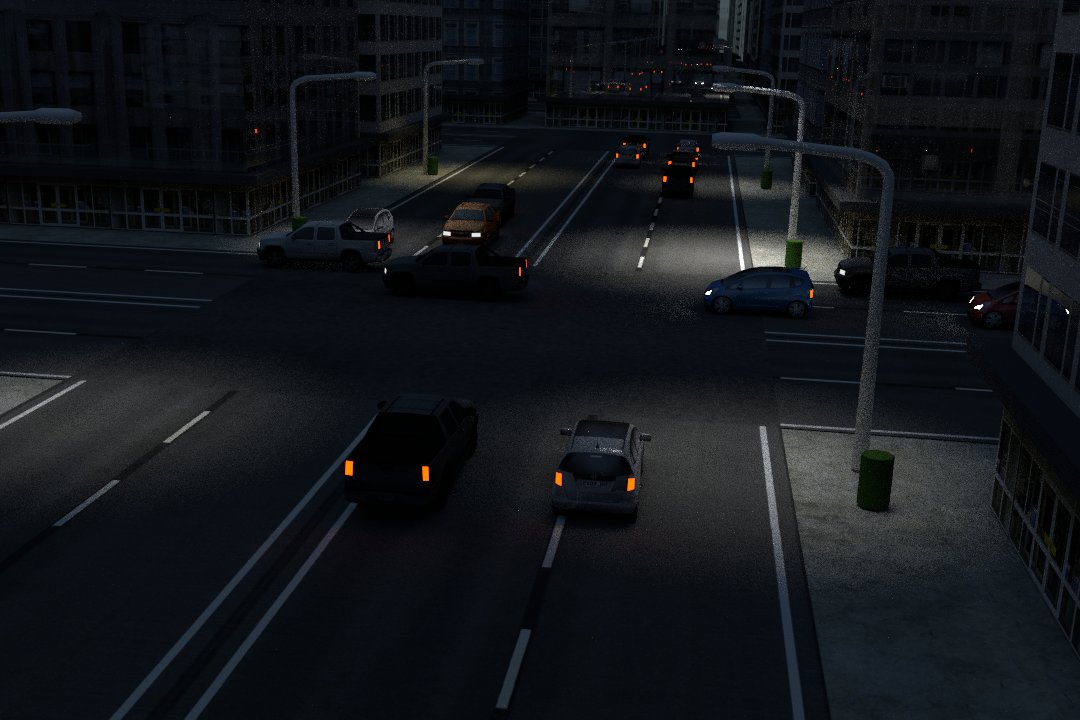}
\includegraphics[width=5.7cm,height=3.8cm]{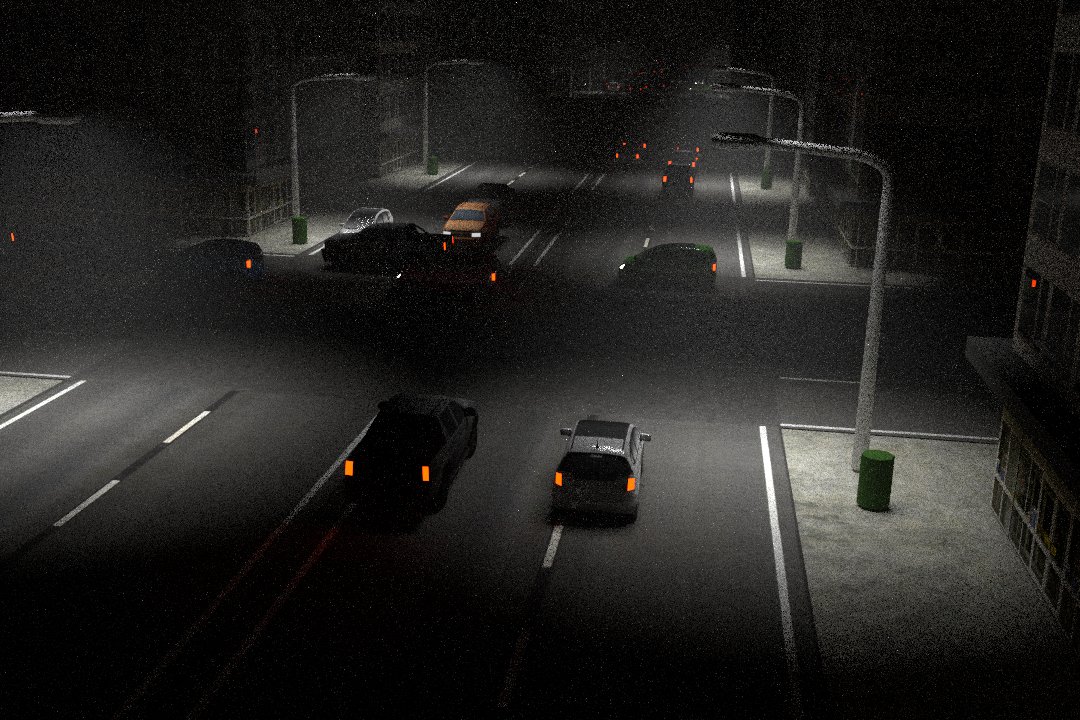}
\caption{\small{Different states of same scene: \textit{Top row (left to right)}:- scene state with ambient light and diffuse surfaces (no shadows), Noon state (with specularities activated), Post-noon;
\textit{Bottom row (left to right)}:- Overcast sky illumination, Night with street lights turned on; and Night with fog.}}
\label{fig_fidelity}
\end{figure*}

We aim to develop a simulation platform which facilitates sampling the contextual parameters ($\theta_W$) from the domain models ($p(\theta_W)$) to create 3D virtual worlds and renders the data along with required groundtruth. This platform is written on top of well known open source graphics rendering framework, called \textit{Blender} \cite{blender}. We have decomposed the generative process of image/video data into a series of sequential sub-processes inspired from real world physics based image formation procedure. We collected several shaders for each subprocess involved in the generation pipeline and integrated them into this coherent platform. All parameters of these shaders and plugins are exposed to the scripting interface of Blender. Please see the supplementary materials for the details of development and limitations of the current platform. 
%The rendering parameters which influences the fidelity of simulated data (such as selection of graphics algorithm ($G$) and their parameters) are denoted as $\theta_G$. For this work, we used MC path tracing algorithm with parameters of 200 rendering passes. 
One of the limitations of the platform is that it can only generate Manhatten scenes by sampling from marked point processes. Manual adjustment is needed for plausible scene configurations. Some of the rendered samples under different settings are shown in the Figure.\ref{fig_fidelity}.

%To show the utility of the framework, we conduct two experiments. In the first experiment, we take a generative model inspired hypothesis (Rank-Order consistency model) and validate it using graphics simulated data. We compare the qualitative and quantitative conclusions against real world data sets. Motivation behind this experiment is two-fold: checking the correctness of our implementation of the platform based on known behaviors of existing models and demonstrating the use of graphics from different levels of inferences ranging from qualitative to quantitative. 

%In the second experiment, we evaluate and compare two modules based on quasi-invariants derived from the Order consistency assumption, for a task of change detection w.r.t. different contexts by modeling their performance.  This demonstrates establishing the empirical mapping from contexts to modules by using physics based simulations. 
%This system engineering philosophy has been advocated by many researchers. 
%In other words, simulations can make explicit the correspondence between physics based variables to invariant behavior. This also demonstrates the fact that the context dependent performance modeling of the alternatives gives hints about further improvement of the module by fusing them to achieve required invariance. These experiments conclude that qualitative inferences and verifications are possible with simulations and illustrates rigorous model based science and engineering that reinforce each other. 

%-------------------------------------------------------------------------

\section{Validating a Hypothesis}\label{sec_hyp_validate}

As we discussed before the degree of insights gained from simulations can provide qualitative and quantitative information depending on the match of fidelity of simulation models to reality. In order to certify conclusions drawn from simulation based testing to be valid on real-world data, one option is detailed estimation of parameters of the attributes input to the physics based graphics engine from real data. This may need to be carefully done in settings where safety critical system requirements are present. However, a middle ground is to take input data, use human experts to postulate stochastic models for scene generation and generate stochastic scenes and images whose statistical properties are similar to real data samples, and then explore a range of physics engine parameter settings to generate a population of data. In this situation, some of the parameters input the simulation may have higher precision, while other parameters are adjusted by human experts to qualitatively match the output characteristics desired.  We pursue this option in this paper. 

More precisely, to assess the qualitative and quantitative conclusions from graphics simulations, we append an hypothesis from generative model based vision literature to the simulation platform and validate it using simulations. Next, we compare insights from the simulation results to that of real world. In this section, we focus on a specific vision model, \textit{Order-consistency} model which hypothesizes that photometric transformations are quasi-monotonic. The qualitative and quantitative conclusions by validating this model using graphics simulations are compared against the real world data.

\subsection{Rank-Order Consistency Model}

Object detection in video surveillance systems is typically achieved through the use of background subtraction or change detection modules. The design of these modules involves modeling quasi-invariant measures which are insensitive to illumination changes and sensitive to geometric changes. A family of invariant operators \cite{singh2008order,xie2004sudden,bhat1998ordinal,zoghlami2009illumination} are derived from \textit{Rank-Order} (RO) consistency assumption, i.e. that photometric transformations can be approximated as locally monotonic. 

%Order based statistics such as Rank-order transform are invariant to monotonic transformations \cite{singh2008order}. Note that it is parametrized by localness (patch size):
%\begin{equation*}
%\Omega^{(s)}(t+\bigtriangleup t) = P\{ \Omega^{(s)}(t) \} , 
%\end{equation*}
%\begin{equation}
%\pi^{(s)}_{t+\bigtriangleup t} = \pi^{(s)}_t \ \ \ \ if\ P\ is\ monotonic .
%\end{equation}
%where $\Omega^{(s)}(t)$ is an image patch of size $s$X$s$ at time $t$ and $P$ is an unknown photometric transformation. $\pi(t)$ is a vector of order indexes of the patch $\Omega(t)$. 

\begin{figure*}[!htb]
 \centering
\subfloat[Validation of RO model under global illumination change \label{fig_model_oc_global}]
   {\includegraphics[width=18cm,height=5cm]{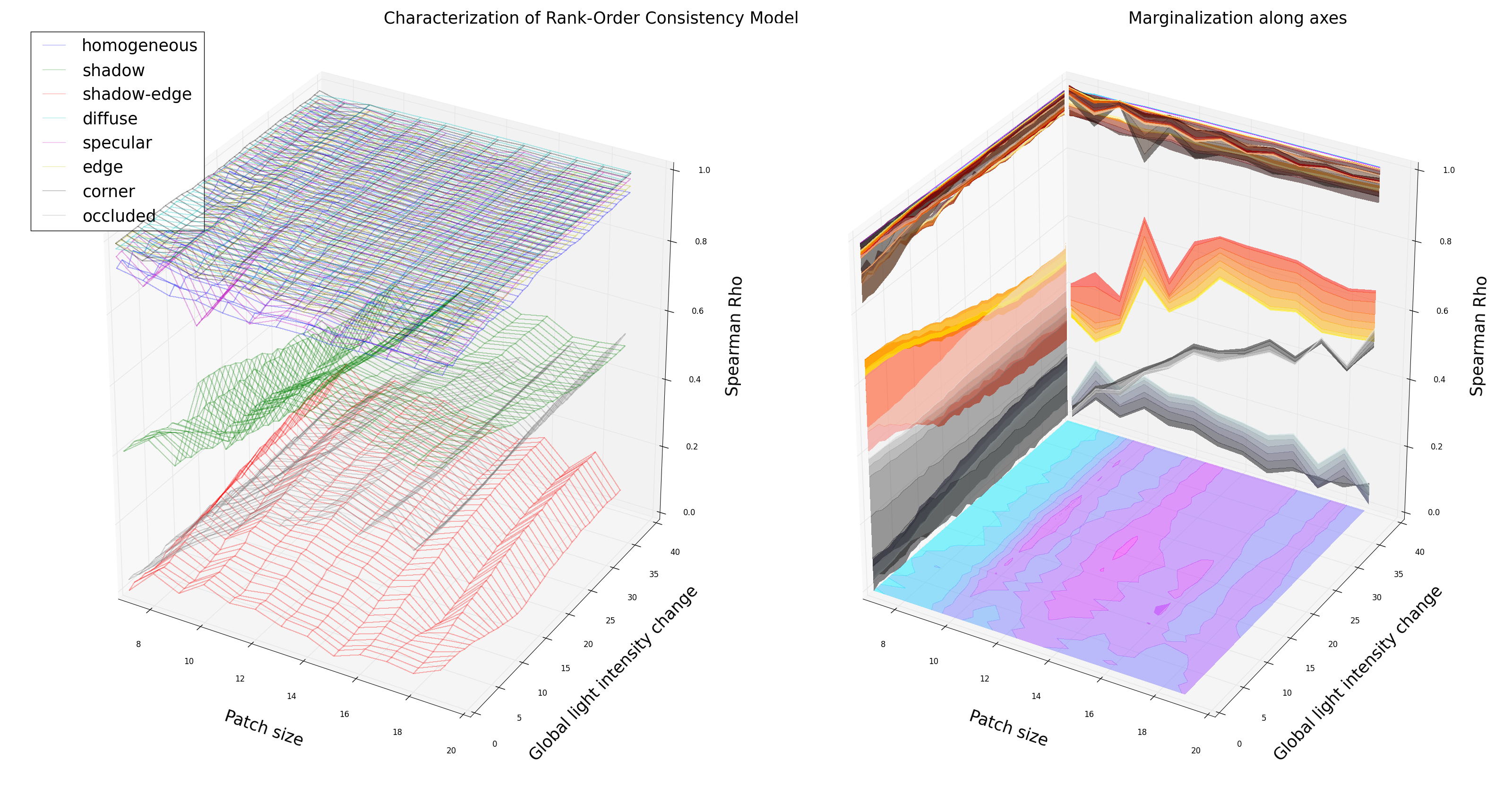}} \\
\subfloat[Validation of RO model under local illumination change \label{fig_model_oc_local}]
    {\includegraphics[width=18cm,height=5cm]{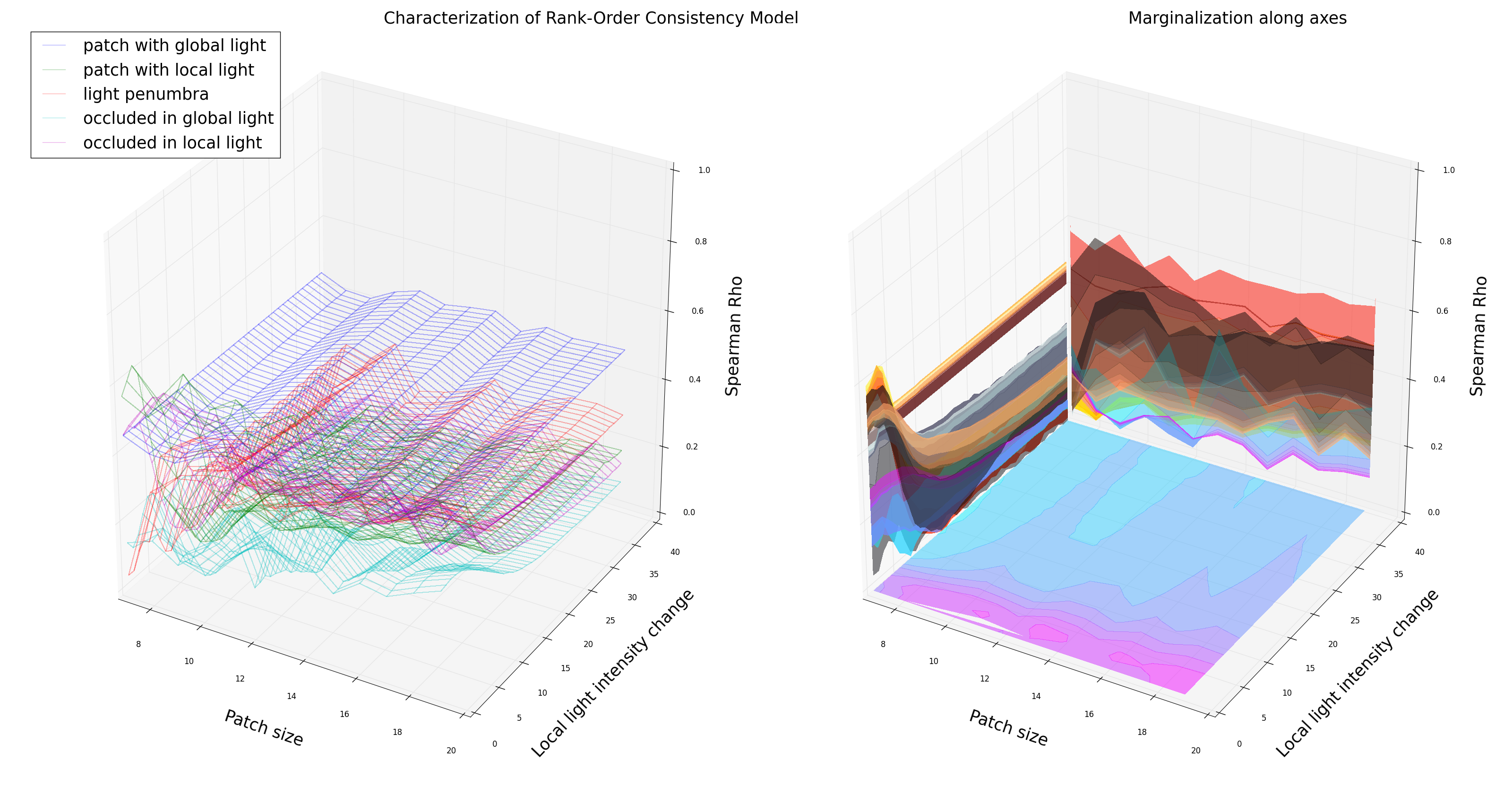}} \\
    \subfloat[Validation of RO model under fog density change \label{fig_model_oc_weather}]
    {\includegraphics[width=18cm,height=5cm]{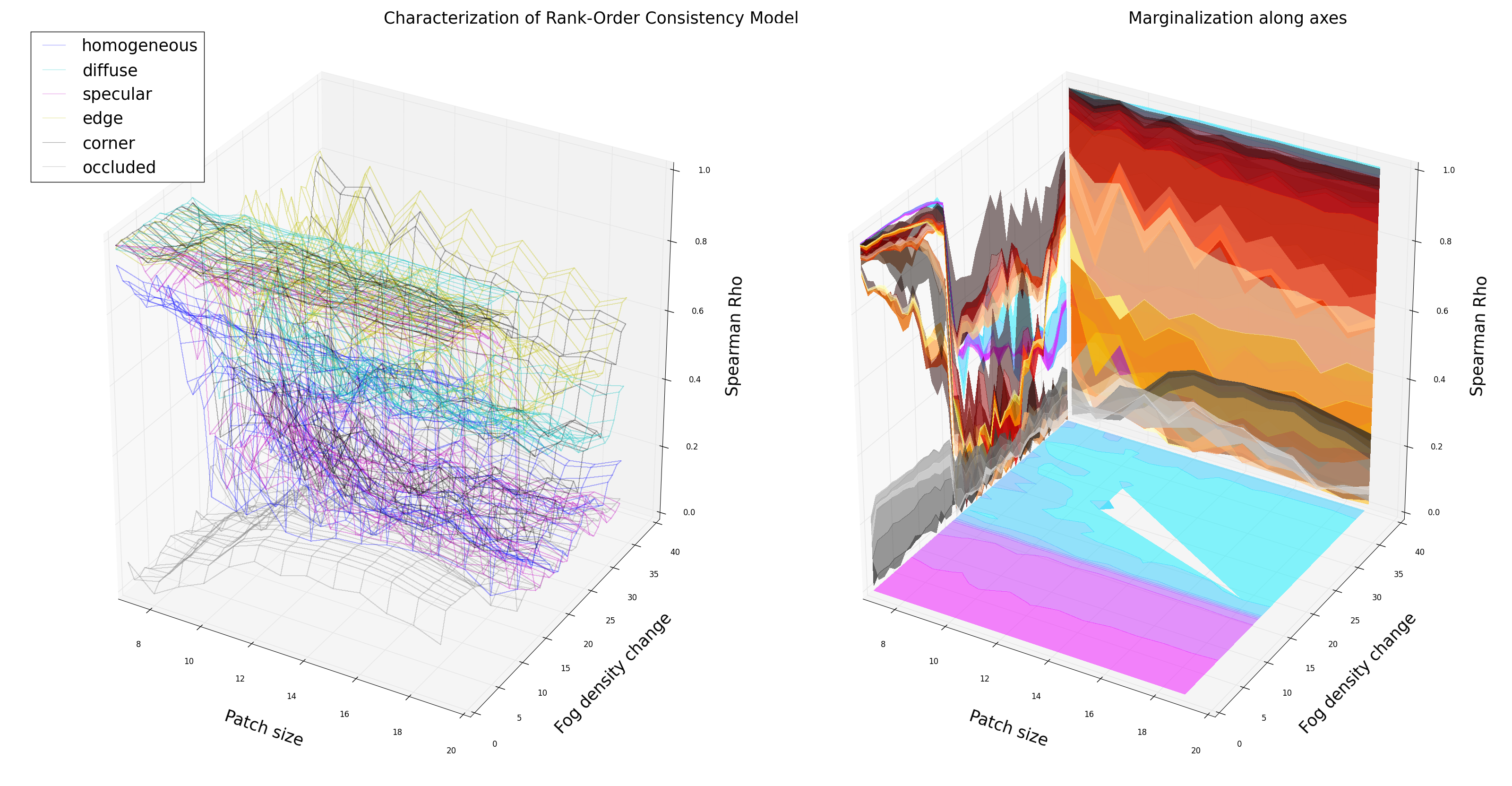}} 
\caption{ \small{We take spearmann's $\rho$ as criterion function to measure the performance of RO model, and plot it as a function of contextual variable and patch size by simulations under different temporal contexts (left plots). Right column plots are mere integration of characteristic manifolds of corresponding left plot, provided for better visual analysis of overall performance.}}
\label{fig_oc_model_validations}
\end{figure*}

We validate this model under different contexts by establish its behavior as a function of context and patch size. We choose Spearman's rank correlation \cite{zar1972significance} coefficient ($\rho$) as a measure of monotonocity of the patch transformation. Other criterion measures can be considered depending on the task and user's interest. From the performance modeling point of view, we can analyze the behavior of this model by establishing criterion measure as a function of its tunable parameters and contextual parameters \cite{ramesh1995performance}:
%\begin{equation}
$\rho = f_{RO}(\theta_W, s)$,
%\end{equation}
where $\theta_W$ is a contextual variable accounts for both spatial and temporal contexts. The nonparametric version of $f_{RO}$ is computed with simulations and shown in the Figure \ref{fig_oc_model_validations}. The details of computation and inferences drawn from it are provided in the later sections.

\begin{table}
\centering
\small
\begin{tabular}{|p{1.7cm}|p{2cm}|c|c|} \hline
Spatial context & Temporal context & $\rho$ (simulated) & $\rho$ (real) \\ \hline \hline
Homogeneous & Global Illumination change & 0.7868 (4) & 0.4457 (5) \\ \hline
Diffuse & Global Illumination change & 0.8323 (2) & 0.5968 (4)\\ \hline
Shadow boundary & Global Illumination change & 0.0877 (6)& 0.6046 (3)\\ \hline
Edge & Global Illumination change & 0.8076 (3)& 0.8313 (1)\\ \hline
Corner & Global Illumination change & 0.8350 (1)& 0.7574 (2)\\ \hline
Occluded  & Global Illumination change & 0.2622 (5) & 0.2635 (6) \\ \hline \hline
All & Day light & 0.6691 (1) & 0.6472 (1) \\ \hline
All & Night & 0.2386 (3)& 0.2550 (3)\\ \hline
All & Fog & 0.4618 (2)&  0.5429 (2)\\ \hline 

\end{tabular}
\caption{\small{Comparison of average of $\rho$ values for different contexts across real and simulated sequences. Numbers in the brackets are ranks.}}
\label{tab_compare_oc}
\end{table}

\subsection{Validation by Simulations}
RO model hypothesizes that the ordering indices of co-located patches should be quasi-consistent given that the change in the patches is only due to photometric transformations. The similarity between order indices can be measured with absolute spearman-rho ($\rho$). $\rho=1$ represents strict order-consistent, while $\rho=0$ represents perfectly non-consistent behavior. We simulated a series of images with same scene configuration but under different temporal contexts of global illumination changes, local light changes, and weather changes. Reference image (scene without dynamic objects) is rendered under ambient illumination conditions. Data generation, rendering models and parameters used for simulations, are more elaborated in the supplementary. For the context of global illumination change, a series of images are rendered with increasing levels of light source intensity to mimic morning to noon sun variations.  Similarly for other temporal changes, images are rendered by increments in corresponding scene parameters. Patches from different spatial contexts (homogeneous region, shadow region, shadow boundaries, diffuse, specular surfaces, edge, corner and occlusions) are sampled in the images. For each spatial context, $\rho$ is computed between patch (sampled from image) and co-located patch in reference image. These values are plotted for different temporal contexts, in Figure \ref{fig_oc_model_validations} (left column plots). 

%%%% draw insights
First, we test the model under a temporal context of global illumination change. The left side plot of Figure \ref{fig_model_oc_global} shows the characteristic manifolds of the models for different spatial contexts under different levels of global illumination increments.  We see that the model is consistently performing better for lambertian (diffuse) surfaces (see the cyan-colored manifold in left side plot of Figure \ref{fig_model_oc_global}). The mean and standard deviation of $\rho$ values on diffuse surfaces are 0.98935 and 0.00423 respectively. Similar behavior is observed on a specular patch whose reflection components were only from background objects (pink-colored manifold). The mean and standard deviation of $\rho$ for this case are  0.95641 and 0.02612 respectively. We also experimented with a specular patch which is illuminated by reflections coming from vehicles (we treat vehicles as foreground objects). These patches behave just like occluded patches for which RO model fails (see gray-colored manifolds). The mean and standard deviation of $\rho$ for this case are  0.29344 and 0.13764 respectively. Observe the mean is too low and standard deviation is too high as the change is due to geometric transformation in the occluded patch. RO model is performing good on high gradient textured patches like edge (yellow) and corner (black) regions. However, the model seems to be failing in the shadow regions. Shadow patch is nearly inconsistent (green-colored manifold) due to some sampling noise of rendering algorithm. This patch might improve its consistent behavior if we allow more rendering time. Shadow boundary patch behaves as order inconsistent (red-colored manifold) due to nonlinear illumination (direct and indirect illuminations). The mean and standard deviation of $\rho$ for the model under this context are  0.10022 and 0.05105 respectively.  To analyze the overall and conditional performances of the model, we integrated the manifolds along the axes and projected on the corresponding planes (see right plot). For the projection on the ground plane, characteristic manifold of occlusion patch is skipped in integration, because it is with geometric change. The model is supposed to be consistent for photometric changes. The marginal variances of model deviations on diffuse, homogeneous patches are less, compared to occluded and shadow boundary patches. The model seems performing relatively better with patch size 13X13 for all levels of increments (magenta region on the ground plane). Hence, optimal patch size for this model is 13X13.  For lower patch sizes, the model is worse (cyan-colored region). We observe that characteristic manifolds for the patches in homogeneous and shadow regions are with relatively high variance due to high frequency sampling noise in numerical (MC path tracer) rendering process.

%%% does these insights match with reality, how do you compare. 
To validate the utility of the graphics for global illumination change simulations for RO model validation, we compare these insights to that of similar real world sequences both quantitatively and qualitatively. We carefully selected the real videos from the benchmarking datasets \cite{goyette2012changedetection}, which capture similar temporal contexts considered above. Assuming first frame of these videos as reference image, we computed $\rho$ values, averaged over several patches sampled from each spatial contexts (similar to the ones considered in the above). These values are provided in Table \ref{tab_compare_oc} along with the ones on simulated data. From the table, it is clear that quantitative performance of the model is quite different across simulated and real world experimental settings. However, we observe that the qualitative statements and ordering of spatial contexts, made on the simulated data, are close to reality to some extent. 
More over, these insights also match with the statements found in the rank-order literature about its behavior on the type of patches \cite{singh2008order,xie2004sudden,bhat1998ordinal,zoghlami2009illumination,mittal2006intensity}. 

We also validated the model under other temporal contexts such as local light changes and weather changes etc. Characteristic manifolds for different spatial contexts, are displayed in Figure \ref{fig_model_oc_local} and \ref{fig_model_oc_weather} for local illumination and fog density change respectively. We can also observe the model behavior (see the topographical surfaces of manifolds of Figure \ref{fig_model_oc_local} and \ref{fig_model_oc_weather}) for the patches that are illuminated by local (street lights or traffic signal lights) or global lights (sun) and patches at light penumbra (nonlinear spatial variations in light) etc. The model is able to preserve its consistency only for first few levels of increments in local light and fog density (see the magenta regions on the ground planes of right plots). $\rho$ values are averaged over all patches (from all spatial contexts) under different temporal contexts are also shown in Table \ref{tab_compare_oc}.  Please see the supplementary for details and real world samples considered for these experiments. However, we observe some deviations in the rankings of spatial contexts under global illumination change (see the ranks in the brackets in Table \ref{tab_compare_oc}), even though ranking of temporal contexts averaged over all spatial contexts is same across real and simulated data. These differences in values of $\rho$ and their rankings is most likely due to mismatch in the contextual models, level of fidelity achieved. The model seems to be best for day light scenes and fails in night scenes and bad weather. These qualitative statements and rankings of contexts are matching across real and simulated worlds. Since the RO model is derived from photometric observations, graphics should simulate light propagation in the scene as accurate as possible (and thus photorealism) to be certified for validation these kind of models and to transfer photometric information to real world systems. Hence, one might give constant concern about correctness of the physics of the graphics for these situations. 
%We also validated other invariant features derived from RO model using graphics to index optimal features to simulate a change detection system for outdoor static surveillance application. These experiments are provided in the supplementary.

\begin{table*}[htb]
\begin{center}
\footnotesize
    \begin{tabular}{ | p{2cm} | p{4cm} | p{4cm} | p{4cm} | } % for centering { | c | c | c | c | }  
    \hline
    $M_0$ & $M_2$ & $M_2 $ & $M3 $ \\ \hline
    SSD & NCC & Ordinal Measure & DCT + Ordinal Measure \\ \hline
    $\sum \limits_{i=1}^n (I_1 ^i - I_2 ^i)^2 $ & $\frac{\sum \limits_{i=1}^n (I_1 ^i - \bar{I_1})(I_2 ^i - \bar{I_2})}{\sqrt{\sum \limits_{i=1}^n(I_1 ^i - \bar{I_1})^2} \sqrt{\sum \limits_{i=1}^n (I_2 ^i - \bar{I_2})^2} }$&  $d_{RO}(\pi_1, \pi_2) \ \ \ \ \ \ $ & $d_{RO}(R_c^{(1)}, R_c^{(2)})$,  $R^{(i)}$ is rank ordered $i^{th}$ patch in DCT-space,  $c=\{1,2, 3, ...k\}$\\ \hline 
    $G = I$ &  $G = I$ &  $G = I$ &  $G = I$ \\  
    $P = I$ &  $P = T, S$ & $P = Monotonic$ & $P = Monotonic$ \\ 
    $N = Gaussian$ & $N = Gaussian$  & $N = $ Symmetric noise distributions   & $ N = $ Replacement noise in Transform space \\ \hline
    \end{tabular}
\end{center}
\caption{\small{Nested model with invariance properties: $G$ corresponds to the type of geometric transformations, $P$ corresponds to the photometric transformations, and $N$ corresponds to the type of perturbations in a given context. 
That is: $G$ corresponds to one of \{ $I$, Translation, Scale, Rotation, Tilt, Pan, Shear \}, $P$ corresponds to one of \{ $I$, Translation, Scale, Monotone, Other \}, 
and $N$ corresponds to one of \{ Gaussian, Replacement noise (salt \& pepper), structured perturbations  \}}.
}
\label{table:NestedModel}
\end{table*}

We observe from the projection plots (Figure \ref{fig_model_oc_global}), that even simply by thresholding on $\rho$ at 0.8 (for a 13X13 patch size) it is possible to obtain a moderately accurate change detector whose output produces false positives largely in shadow regions.   However, this model can be combined with discrete cosine transform (DCT) to expand the model to be quasi-invariant to high frequency noise \cite{zoghlami2009illumination}. We will also validate this compound quasi-invariant model in the next section to design a change detection system using simulations.  

\section{Experiments with Graphics Simulated Patch Population} \label{sec_changedetect}
In this section, we illustrate that the  graphics simulation tools, combined with model based thinking, provides complementary insights and provide exploration ground for devising new models to achieve required invariance to other perturbations.  We also illustrate the use of simulation tools along with algorithms that have generative model based semantics but no physics-based model correspondence.
%In this section we illustrate the use of simulation tools along with algorithms that have generative model based semantics but no physics-based model correspondence. %We do this in the explicit context of change detection in video. While these experiments help link physics based model parameters to algorithm behavior (in black box sense), we also illustrate that model based thinking provides complementary insights and provide exploration ground for devising new physics based models. 
We first provide a classification of various modules for patch matching and how they correspond to specific generative modeling assumptions.  The models in general do not form a strict nested ordering but involve partial ordering based in terms of the signal model and perturbation model parameters. Specifically, two block level change detection schemes, Rank-Order (RO) based matching in gray scale space and rank order matching in Discrete Cosine Transform (DCT) space, are compared in this section using the graphics simulations to illustrate their level of invariance to monotone transformations and structured perturbations.  Graphics simulations involving environmental states such as fog and rain correspond to structured perturbations while illumination changes correspond to monotone mappings. We use a population of patch data constructed using these simulations and added perturbations according to classical methods for performance characterization. 
%The motivations for use of DCT, RO based representations for change detection are elaborated in the supplemental material.
%Intensity functions corresponding to patches can be viewed as partial ordered models involving simple to more and more complex characteristics. 
%Let $G$ correspond to the type of geometric transformations, $P$ correspond to the photometric transformations, and $N$ correspond to the type of perturbations in a given context. That is:
%$G$ corresponds to one of \{ $I$, Translation, Scale, Rotation, Tilt, Pan, Shear \}, 
%$P$ corresponds to one of \{ $I$, Translation, Scale, Monotone, Other \}, and
%$N$ corresponds to one of \{ Gaussian, Replacement noise (salt & pepper), structured perturbations  \}. 
Distance functions used in computer vision for comparing patches such as sum-of-squared differences (SSD), Normalized cross-correlation (NCC), Ordinal distances, etc. can be viewed as appropriate depending on the specific contextual setting (i.e. specific choice of $G, P$ and $N$. See the Table.\ref{table:NestedModel}). Thus, a given combination, $(G, P, N)$, corresponds to appropriate module choice for distance computation. 
\textcolor[rgb]{0,0,0}{The Table.\ref{table:NestedModel} summarizes the  properties for various modules and their respective model assumptions. 
} For instance, in the special case of DCT followed by rank based distance computations, one can view the
distance as being quasi-invariant to both offset, scale, monotone transformations and to moderate levels of structured perturbations where the frequency content of the dominant texture is still retained. See for instance:\cite{zoghlami2009illumination}, for the use of a robust distance metric that considers only the contrasting pair of DCT coefficients with largest positive and negative coefficients as basis for representation of the dominant structure present in a pattern. This distance metric can be modified to allow $k$ pairs of DCT coefficients instead of just the pair with significant contrast in coefficient values. There is no obvious and direct correspondence of this distance metric to a published physics based model.  However, the use of graphics simulations with alternative physics based models allows one to establish the correspondence between physics based models to this module. The insight from our experiments can be a useful trigger for physics based vision researchers to think about generalizing various models in their arsenal.

%In our experiments, we study the behavior of two specific modules (one that assumes that ordering of gray values are preserved between two images and another that verifies that the dominant frequencies in transform space do not change between two images ). The motivations for use of DCT, Rank-order based representations for change detection are elaborated in the supplemental material.

% \begin{figure}[hbt]
% \centering
% \subfloat[Gaussian\label{fig:test1}]
%   {\includegraphics[scale=0.25]{images/DCTROResidualNoiseTexture_Gaussian_Simn.png}} \hspace{1mm}
% \subfloat[Gaussian\label{fig:test4}]
%    {\includegraphics[scale=0.25]{images/DCT+ROResidualNoiseTexture_Gaussian_Simn}} \\ 

%\subfloat[Salt n Pepper\label{fig:test2}]
%    {\includegraphics[scale=0.25]{images/DCTROResidualNoiseTexture_SaltnPepper_Simn.png}} \hspace{1mm}
%\subfloat[Salt n Pepper\label{fig:test}]
%   {\includegraphics[scale=0.25]{images/DCT+ROResidualNoiseTexture_SaltnPepper_Simn.png}} \\
    
% \subfloat[Speckle\label{fig:test3}]
%    {\includegraphics[scale=0.25]{images/DCTROResidualNoiseTexture_Speckle_Simn.png}} \hspace{1mm} 
% \subfloat[Speckle\label{fig:test6}]
%    {\includegraphics[scale=0.25]{images/DCT+ROResidualNoiseTexture_Speckle_Simn.png}} 
% \caption{\small{DCT and RO robustness w.r.t Noise and  Texture - Simulated Data, Plots: DCT in green, RO in red and DCT+RO
% in blue.}}%, Illumination and blur }
% \label{DCTRORobustness_simulation}
% \end{figure}
 
\begin{figure}[hbt]
 \centering
\subfloat[Gaussian\label{fig:test1}]
  {\includegraphics[scale=0.25]{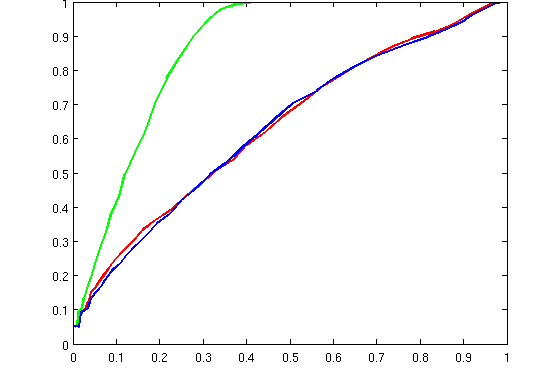}} \hspace{1mm}
 \subfloat[Gaussian\label{fig:test4}]
    {\includegraphics[scale=0.25]{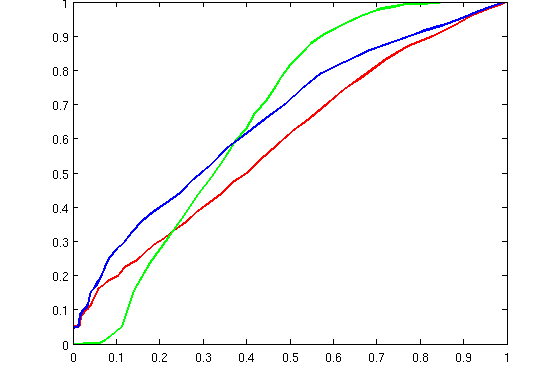}} \\   
\subfloat[Salt n Pepper\label{fig:test2}]
    {\includegraphics[scale=0.25]{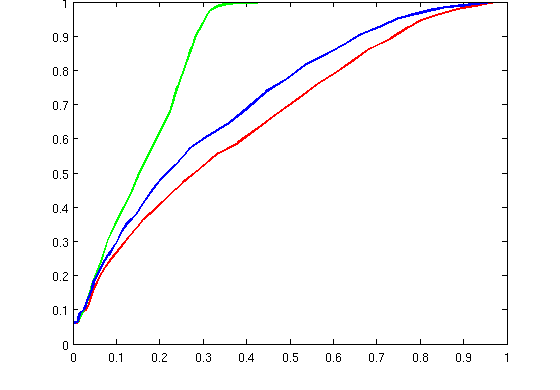}} \hspace{1mm}
 \subfloat[Salt n Pepper\label{fig:test}]
   {\includegraphics[scale=0.25]{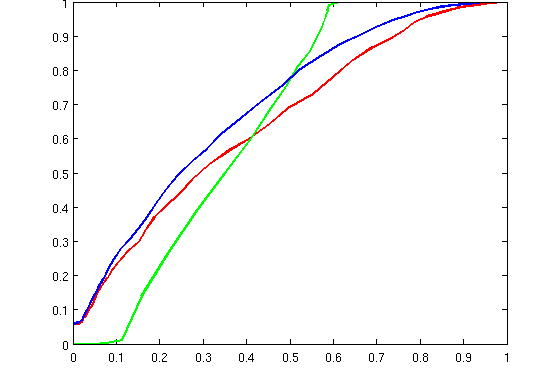}} \\    
 \subfloat[Speckle\label{fig:test3}]
    {\includegraphics[scale=0.25]{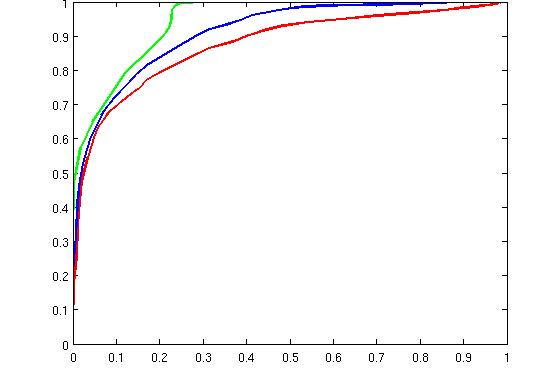}}\hspace{1mm}
 \subfloat[Speckle\label{fig:test6}]
    {\includegraphics[scale=0.25]{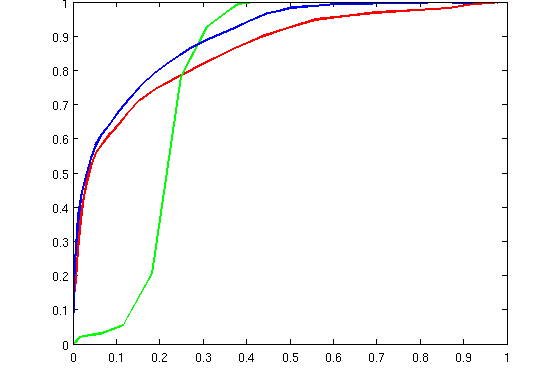}} 
 \caption{\small{DCT and RO robustness w.r.t Noise and Illumination, Plots: DCT in green, RO in red and DCT+RO in blue}}%, Illumination and blur }
 \label{DCTRORobustness}
 \end{figure}

{\bf Robustness w.r.t Noise and Illumination -} To study the robustness of Rank Order and DCT, we analyzed their response with respect to different noise levels and illuminations on different textured patches collected randomly from a pool of images. The experiments are conducted on random image patches and background image patches are perturbed with noise and illuminations. We considered three different types of noise: Gaussian, salt \& pepper and speckle noise. The ROC curves with respect to DCT, RO and their combination is shown in Figure \ref{DCTRORobustness}.  In these figures, the RO response is presented in red color, while green represents DCT measures and their combination (DCT+RO) is showing in blue. In all the three plots (first column) it can be observed that DCT responses are quite high as compared to RO. In case of high frequency noises( salt \& pepper and speckle noise) the DCT response is quite good. RO response is higher in case of speckle noise compared to Gaussian and salt and pepper noise. 
In second column, we present ROC plots from samples with added noises and illuminations.  It can be observed that with illumination perturbations, the DCT performance degrades and performs poorly compared to RO, and the combination DCT with RO performs better in all the three cases. This validates the invariance property of the combined operation. 
We applied the combination (DCT+RO) of operations for change detection in illumination data and results are shown in supplementary material. The effect of noise and illumination are quite visible in the obtained results while applying DCT and RO separately, while the combined operation produces better results.  

%We applied the combination (DCT plus RO) of operations for change detection in illumination data and results are shown  Figure \ref{fig_changedetection}. In the first row we have shown change detection results from RO, DCT result is shown in second row and in the third row the detection of the combined operation (DCT+RO) is shown. The effect of noise and illumination are quite visible in the obtained results while applying DCT and RO separately, while the combined operation produces better result in the considered case. 

\section{Conclusion and Future work}\label{sec_conc}
In this article, we illustrated how a graphics simulation platform's usefulness can be evaluated by thinking of it as a parametrized  system and the deviations in rendered data propagate through the subsequent stage (vision module) to produce 
deviations in the final output.  The significance of the impact of these deviations on the experimental conclusion depends on  the nature of the conclusion drawn that may range from  qualitative to quantitative aspects of a vision system depending on the degree of fidelity and closeness of the simulation models to reality. A physics based graphics platform has been discussed which aims to provide a flexible platform for learning and to aid in exploration of design trade-offs for a range of vision solutions. 
We demonstrated the utility of the platform to provide qualitative to quantitative assessments of performance of models and modules. We also provided a case study in which the simulation platform was to used to establish the link between alternative viewpoints, involving models with physics based semantics and signal and perturbation semantics. The net result is the confirmation of insights for robust invariant change detection \cite{zoghlami2009illumination}. Ongoing and future work involves broader and systematic studies to demonstrate the utility of simulations in quantitative assessment of vision systems. 

%complementary insights and exploration ground for devising new models to achieve required invariance to other perturbations, by establishing explicit mapping between physics based contextual models and algorithms that have generative model-based semantics but lacking correspondence to physics-based models. 
%two specific case studies. One case study validated the Order-consistency model in outdoor scene contexts, while another case study highlighted how the evaluation of algorithms that have generative model-based semantics but lacking correspondence to physics-based models can be achieved by use of the simulation platform. 

\section*{Acknowledgments}
This work was supported by the German Federal Ministry of Education and Research (BMBF) in
the projects, 01GQ0840 and 01GQ0841 (BFNT Frankfurt).

{\small
\bibliographystyle{ieee}
\bibliography{egbib}
}

\appendix
\section*{Appendix}
The Appendix is organized as follows. In Section \ref{sec_platform_details}, we provide the details of the graphics simulation platform which is used to generate the
data for the experiments, discussed in the main article. Section \ref{sec_data_generation} discusses about the data preparation for validation experiments. More elaborated details of the change detection experiments included in the main article, are provided in Section \ref{sec_change_detection}.

\begin{figure*}[!htp]
\centering
\includegraphics[width=12cm, height=4cm]{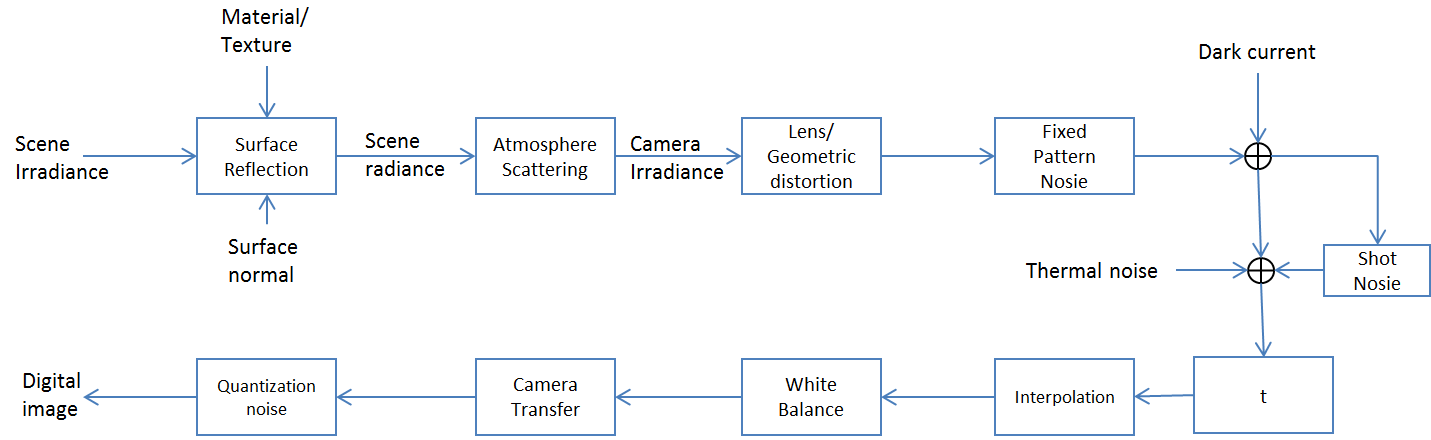}
\includegraphics[width=5cm]{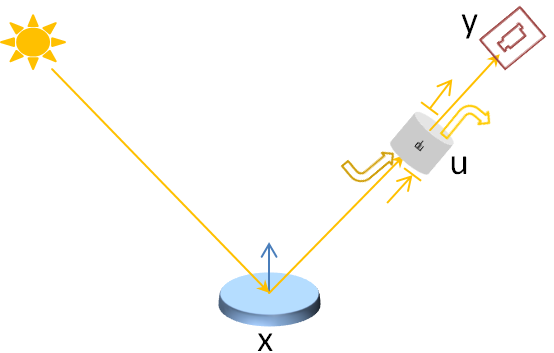}
\caption{\small Light propagation: (a) Physics based decomposition of image generation pipeline (b) Light propagation for image generation}
\label{fig_image_generation_pipeline}
\end{figure*}

\section{Graphics Simulation Platform} \label{sec_platform_details}

 The appearance of a pixel on the image plane is a result of many physical generative processes. The goal of computer graphics is to synthesize image measurements given the description of world parameters according to physics based image formation principles (forward inference) while the focus of computer vision is to map the pixel measurements to 3D scene parameters and semantics (inverse inference). Apparently their goals are complementary, but fields are seeming to converging to a common point i.e physics based models. Our goal is to build a forward generative system by sequential decomposition of the processes involved in image rendering, shown in Figure.\ref{fig_image_generation_pipeline}. There are some domain-specific-simulators \cite{espie2005torcs,lundgren2006evaluation} already available but they rarely consider weather and sensor perturbations. We think that without considering them, one can not validate the graphics for vision. Moreover, realism is achieved in graphics by using lot of scene-specific heuristics and they are adhoc. Integrating them into unified framework in an harmonized manner is a challenging task.

Several graphics platforms are already available and we have investigated different options for the design choices including virtual reality tools, Mitsubha rendering platform  \cite{jakob2010mitsuba} and Blender \cite{blender} etc. Finally, we considered an open source rendering software, \textit{Blender}, as the base for our framework as it provides a collection of different reflection (\textit{BRDF}), transmission (\textit{BTF}), scattering (\textit{BSDF}) models and advanced rendering algorithms. The reasons for this design choice are that: (a) full code access facilitates the altering and adding components at any stage of simulating process, (b) underlying python language makes it easy to incorporate probabilistic and machine learning packages for stochastic scene generation, (c) Blender facilitates easy integration of third party plugins for importing 3D meshes for online repositories and a feature (\textit{RenderLayers}) which helps for rendering the images/videos in multiple (sub)modalities (groundtruth information such as depth, surface normals, semantic labels and so on).

Our framework is organized into a platform of mostly independent components inspiring from sequential pipeline structure of the Figure.\ref{fig_image_generation_pipeline}.a. Then the simulation process will be done through interactions between these components. Because each component has fixed and well defined interface, this framework allows the switching between different implementations for any particular components without effecting remaining parts of the system and makes it is possible to plug and play with different sub-components.

\textbf{Scene management}:  This component is to script and manage the extrinsic properties of the 3D scene entities (such as Buildings, Vehicles etc.), Light sources, Cameras in the virtual world. The scripting interface exploits the features of \textit{Blender} such  as \textit{Graphs}, \textit{Drivers}, \textit{Shaders} and \textit{Nodes} etc. We have collected a several 3D object mesh models and high quality textures from the online 3D repositories (Google 3D ware-house) for each object category. Stochastic generation of 3D scene by sampling from given/learnt contextual models, is also one of our interests.  Currently, this is done by using conventional and blender (\textit{bpy}) python packages etc. Several BRDF and light models have already been available with Blender. We also implement a volume scattering and absorption shader to render the effects of atmosphere in the image. 

%The variables that this component exposes to the user (indirectly through scene scripting) and supplies to other components, are $x, w_i, w_o, \theta_i$

\textbf{Light propagation} : Once scene scripted, the physical phenomena involved in the light propagation and should simulated are, as shown in the Figure.\ref{fig_image_generation_pipeline}.b: Transformation of (a)scene irradiance to scene radiance (at surfaces), which is explained by so called \textit{Global illumination equation} (Eq.\ref{eq_global_illumination}); (b) scene radiance to camera irradiance (through atmosphere or participating media), which is formulated as \textit{Light transport equation} (Eq.\ref{eq_integral_light_transport}); and (c) camera irradiance to image intensity, which is determined by \textit{camera response} function (Eq.\ref{eq_camera_transfer}). We provide final equations here, from the graphics literature for the completeness of the document. For detailed derivations and explanations, please refer \cite{cerezo2005survey}.

\begin{figure*}
\centering
\includegraphics[width=4.27cm]{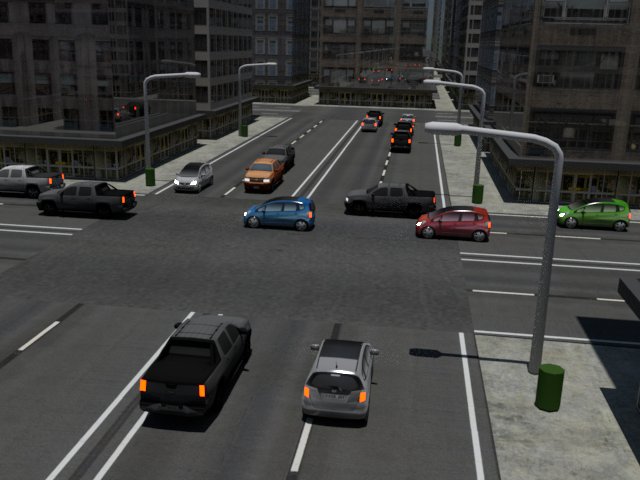}
\includegraphics[width=4.27cm]{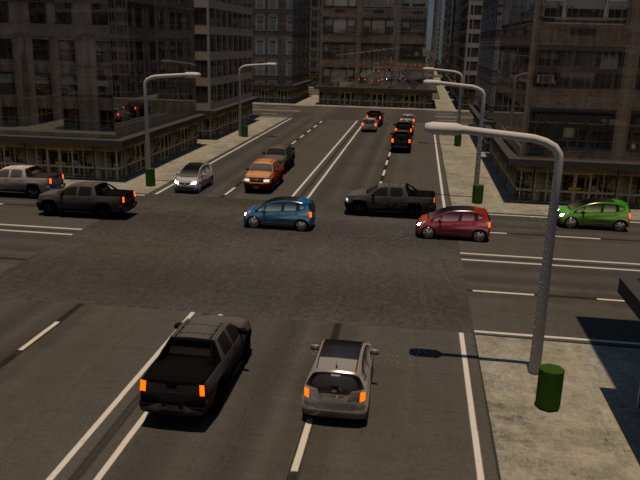}
\includegraphics[width=4.27cm]{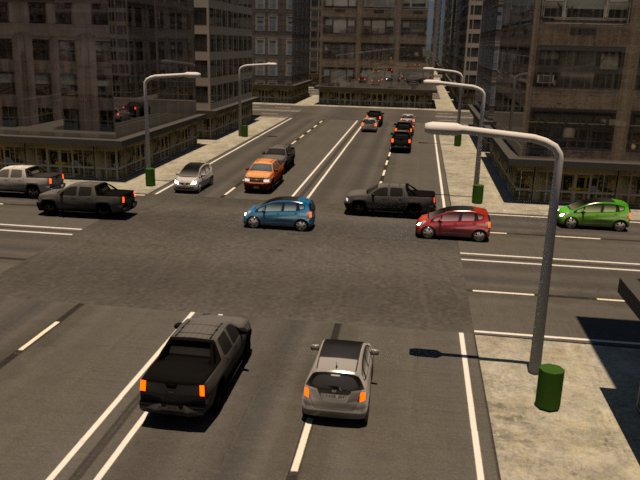}
\includegraphics[width=4.27cm]{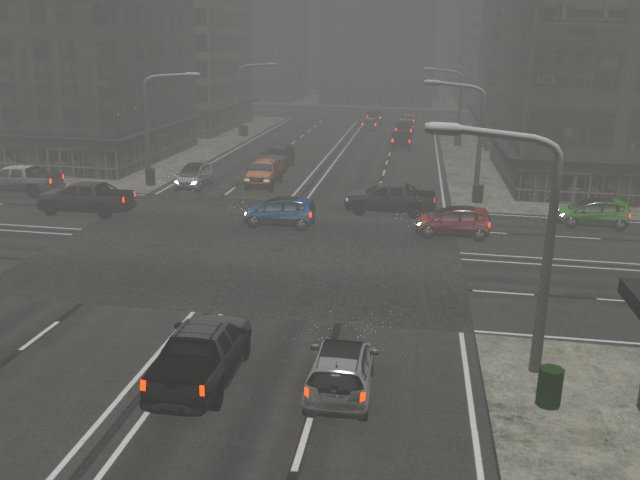}
\includegraphics[width=4.27cm]{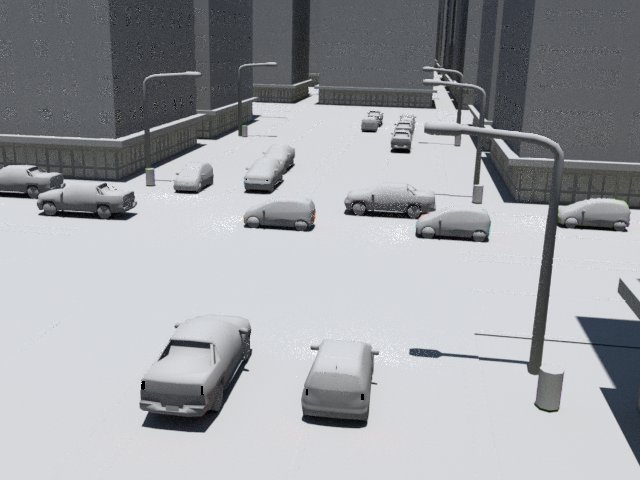}
\includegraphics[width=4.27cm]{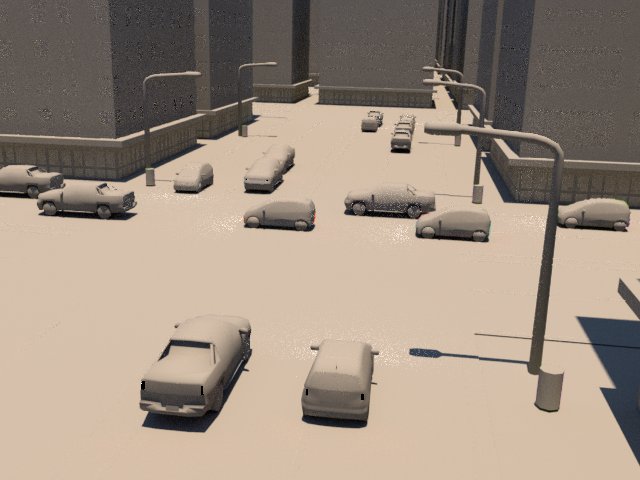}
\includegraphics[width=4.27cm]{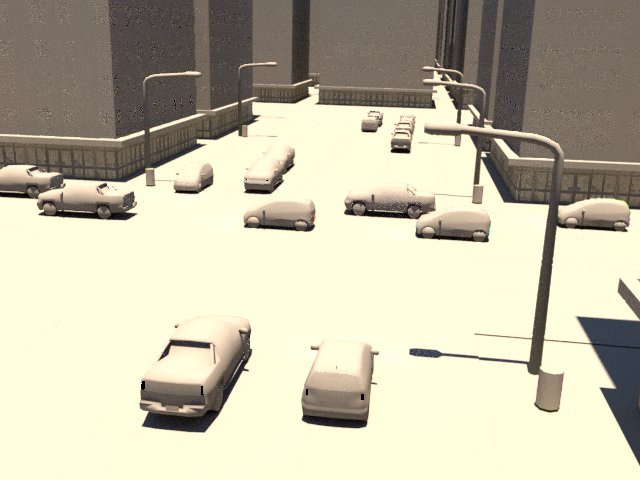}
\includegraphics[width=4.27cm]{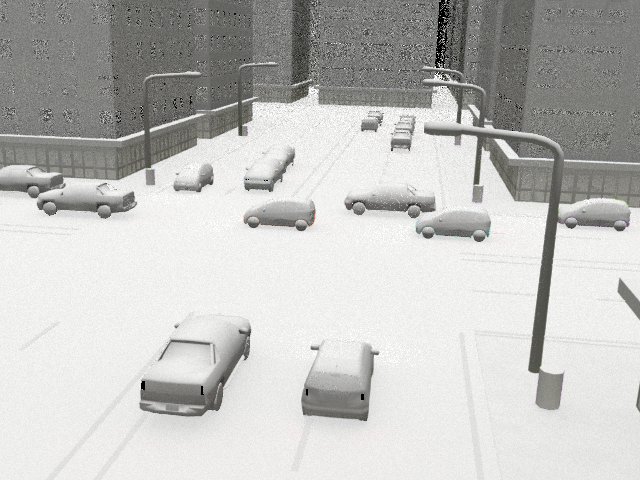}
\caption{\small Variations in \textit{Light source} parameters: \textit{Top row (left to right)} white illuminant, yellowish illuminant, increased intensity levels of the illuminant, fog (volume) scattering effects; \textit{Bottom row} respective shading components.}
\label{fig_light}
\end{figure*}

The global illumination equation is given by, 
\begin{equation}
L_r(x, \omega_o) = L_e(x, \omega_o) + \int_H \rho_{bd}(x, \omega_o, \omega_i) L_i(x, \omega_i) \cos{\theta_i} d \sigma_{\omega_i}
\label{eq_global_illumination}
\end{equation}

where $L_r(x, \omega_o)$ and $L_e(x, \omega_o)$ are the reflected and emitted light in the direction $\omega_o$ at a surface point $x$ $ L_i(x, \omega_i)$ is incident light which has to integrated over a hemisphere $H$. $\rho_{bd}$ is BRDF or BTF, a function of incident ($\omega_i$) and outgoing ($\omega_o$) directions. $\theta_i$ is the angle between incident light ray and surface normal. Physics inspired BRDF/BTF models for diffuse, specular and glassy surfaces are available with Blender. For a good review on BRDF models, please refer to \cite{montes2012overview}. Please note that wavelength dependency in the equations has been neglected for convenience. Emission models ($ L_e$) are used for some purposes in our simulations, for example, brake-lights of vehicles. The effects such as inter-reflections, shadows are generated as a result of appropriately solving the equations. In general, rendering is the process which solves  Eq.\ref{eq_global_illumination} and Eq.\ref{eq_integral_light_transport} simultaneously, either analytically or numerically. Numerical methods are more popular such as Monte carlo path tracing (named as \textit{Cycles} in Blender), Metropolis light transport and Hierarchical radiosity etc. Figure \ref{fig_light}.a-c are the examples of MC path tracer under different lighting configurations. Corresponding shading components are depicted in the bottom row.

Before the scene/surface radiance ($L_r$) transforms to camera irradiance, it has to travel through medium/atmosphere. In this phase, it undergoes three kinds of phenomena: absorption, scattering and emission (see Figure.\ref{fig_image_generation_pipeline}.b) \cite{cerezo2005survey}. These are formulated by the integral light transport equation, $E(y)$ =

\begin{equation}
\small
\begin{split}
\underbrace{  L_r(x, \omega_o) e^{-\int_x^y k_t(u)du} }_{\text{ $E_{al}$: attenuated light}} + \underbrace{\int_x^{y} k_a(u) L_e(u)   e^{-\int_u^y k_t(u)du} du  }_{\text{$E_{el}$: emitted light}}  \\ 
+  \underbrace{  \int_x^y  \int_S \frac{k_s(u)}{4\pi} e^{\int_u^y k_t(u)du}   L_r(x,\omega_i) p(\omega_o,\omega_i) d\sigma_{\omega_i} du }_{\text{$E_{ss}$: single scattering}}\\
+  \underbrace{  \int_x^y  \int_S \frac{k_s(u)}{4\pi} e^{\int_u^y k_t(u)du}   L_m(u,\omega_i) p(\omega_o,\omega_i) d\sigma_{\omega_i} du }_{\text{$E_{ms}$: multiple scattering or medium radiance}}
\end{split}
\label{eq_integral_light_transport}
\end{equation}

where $k_a$, $k_t$ and $k_s$ are absorption, extinction and scattering coefficients of medium. $L_m(u)$ is medium radiance due to multiple scattering and in integrated over sphere $S$ and along the light travel. Due to emission and inscattering, radiance increases because of light impinging on a point $u$ that is scattered into the viewing direction. The spatial distribution of the scattered light is modeled by the phase function $p(\omega_o, \omega_i)$ and different phase functions (such as Mie) have been proposed and applied to simulated polluted sky, haze, clouds, and fog etc. In this work, we use Schlick phase functions \cite{jarosz2008efficient} that are parameterized by anisotropy and particle density. These are proven to be well approximations for theoretical functions and well suited for Monte Carlo rendering methods. %Schlick phase functions for aerosols (isotropic) and haze (anisotropic) are shown in the Fig.\ref{fig_schlick_phases}. 
Please see Figure \ref{fig_light}.d and Figure \ref{fig_bad_weather_sim} for  scenes with fog weather.  To create dynamic weather conditions such as rainy and snow scenes, we use particle random processes \cite{reeves1983particle} with water droplets or snowflakes as particles. Please note that our current process of dynamic weather simulations may not be physically accurate as it changes surface characteristics (rain makes them wet) and appearance depends on capture time (creates blurring effect) \cite{garg2004detection}. These effects can be modelled with \textit{DynamicPaint} feature in Blender and would be considered for future work.
%
%\begin{figure}[h]
%\centering
%\includegraphics[scale=0.2]{images/schlick_rayliegh.png}
%\includegraphics[scale=0.2]{images/schlick_haze.png}
%\caption{Schlick phase functions of $p(\omega_o,\omega_i)$ for Aerosols and Haze particles \cite{jarosz2008efficient}}
%\label{fig_schlick_phases}
%\end{figure}

The goal of rendering algorithms is the resolution of the integral transport equation and global illumination equation, atleast for the points and directions in the view frustrum. The \textbf{Render} component in our framework is supposed to have rendering methods (including traditional and advanced). The current implementations with in the framework are: Path tracing, Metropolis light transport, Ray tracing and Hierarchical Radiosity. We just collected all of them and integrated them into a single platform to facilitate easy plug and play experimentation. The variables that this components exposes to the user (scripting interface) and supplies to other components, are the method to solve the transport equations and its (free/tunable) parameters such as samples, iterations etc. 
 \begin{figure*}[ht]
\centering
\includegraphics[width=4.27cm]{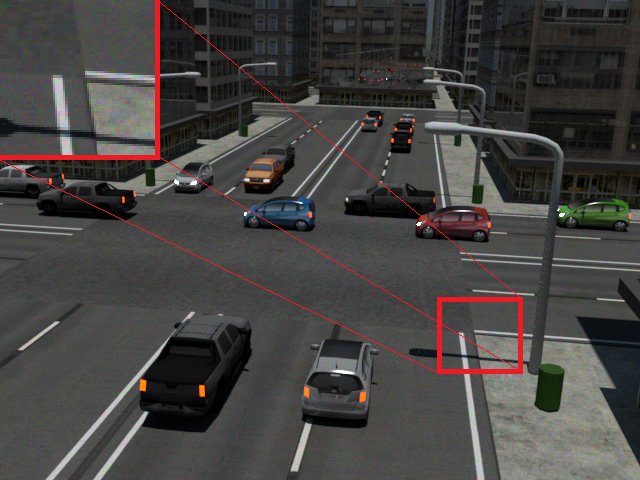}
\includegraphics[width=4.27cm]{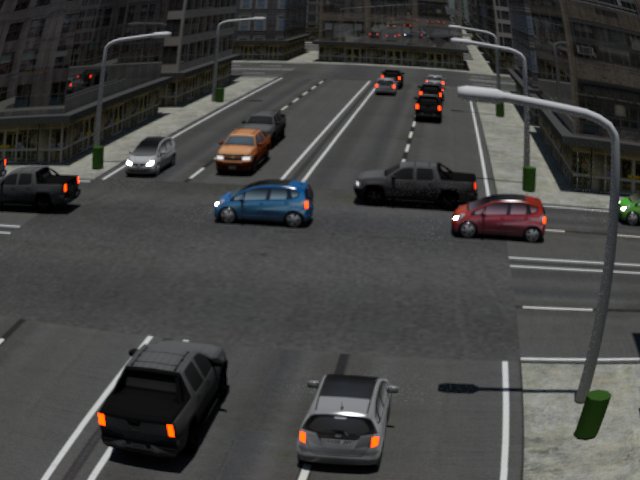}
\includegraphics[width=4.27cm]{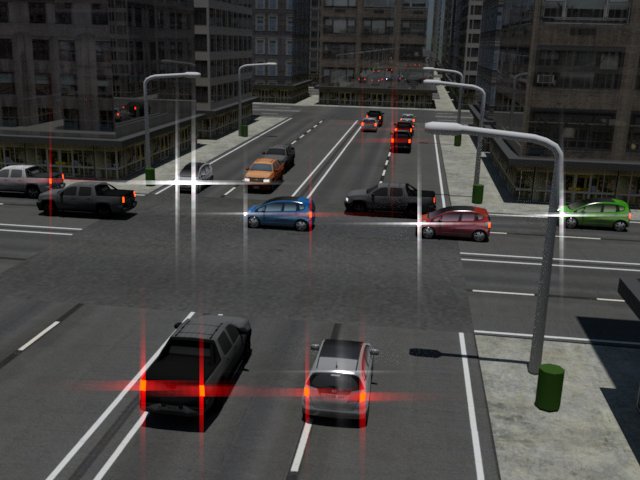}
\includegraphics[width=4.27cm]{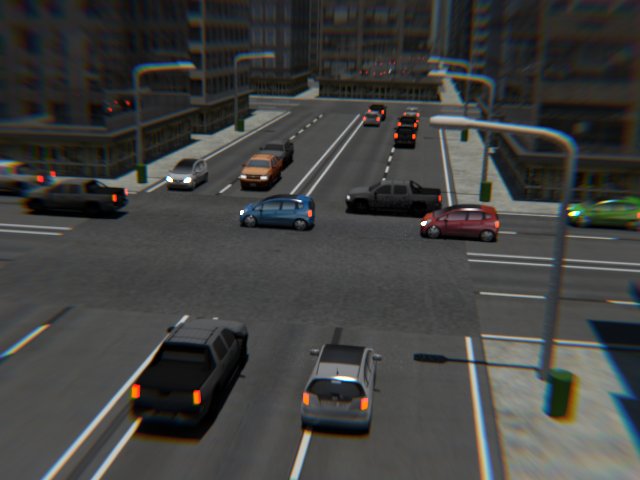}
\caption{\small simulation of sensor effects, \textit{left to right}: Jitter noise (see highlighted patch), Lens distortions, camera glare, chromatic aberrations. }
\label{fig_sensor}
\end{figure*}

\textbf{Sensor}: Computer graphics domain's major focus is about photo-realistic rendering methods and they rarely consider sensor models and their influences on the rendered images. Even if the simulated images are very photo-realistic, they might not be well suited for machine vision applications unless camera imperfections are considered, which include chromatic aberration, vignetting, lens distortion, sensor noises etc. 
Hence, a physics inspired sensor model \cite{tsin2001statistical} has been implemented using \textit{OpenGL} pinhole camera and noise models and some of the effects are done using postprocessing shaders on the \textit{framebuffer}. According to the model, the digitized pixel is given by,

\begin{equation}
\hat{z} = f(a \int_0^t E(y) dt + N_s + N_{c_1} + b) + N_{c_2}
\label{eq_camera_transfer}
\end{equation}
where $\hat{z}$ is the image intensity, $f$ is the camera response function; $a$ and $b$ are the parameters of while balance module which linearly stretches the received camera irradiance $E(y)$ in the exposure time $t$. $N_s$, $N_{c1}$ and $N_{c2}$ are factors due to shot noise, thermal noise and quantization noise. We also provide some post-processing shaders for lens distortions, chromatic aberrations, camera glare effects etc. Some sample results for camera effects are shown in Figure \ref{fig_sensor}. 
%For the work in the later sections, we use  the sensor settings: $b=0$, $f$ is linear with different slopes for RGB channels as in \cite{narasimhan2003models}, $variance(N_s)=0.015$, $variance(N_{c_1})=0.01$ and $variance(N_{c_2})=1$. Some of the rendered samples under different settings are shown in the Fig.\ref{fig_fidelity}.

\begin{figure*}
\includegraphics[width=4.27cm]{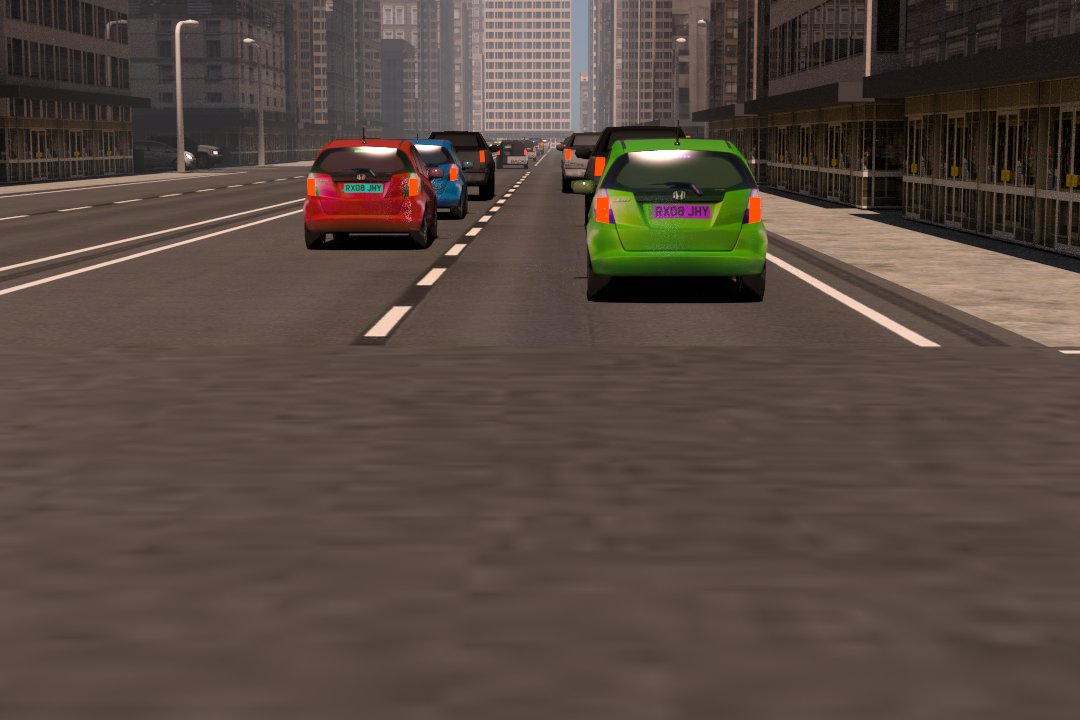}
\includegraphics[width=4.27cm]{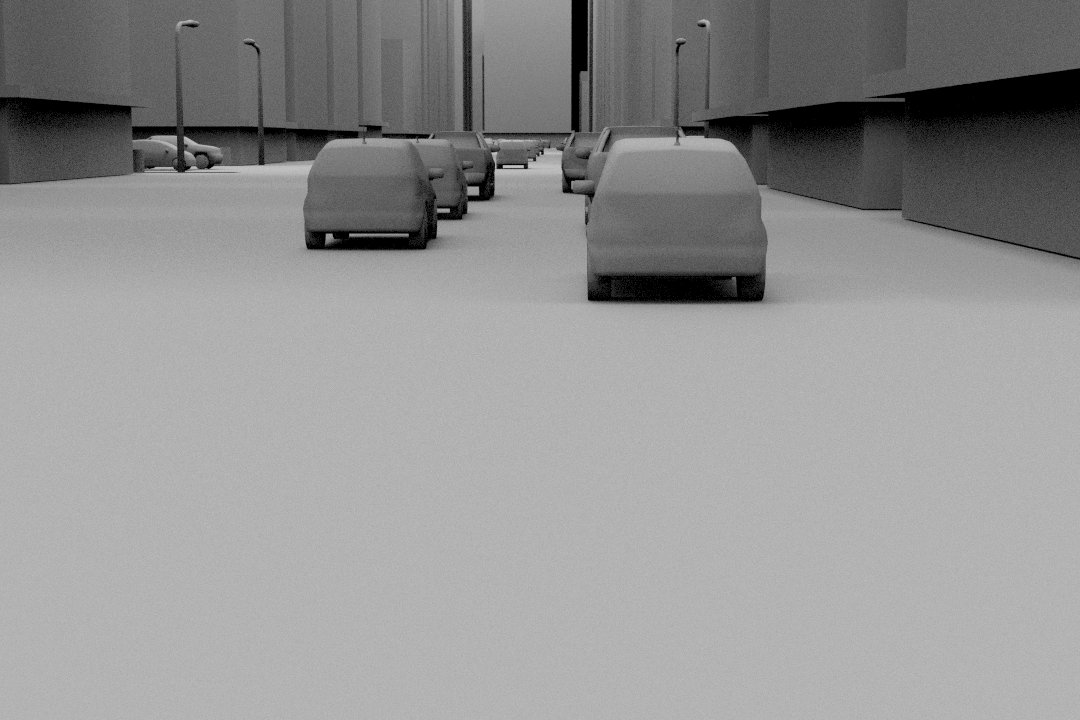}
\includegraphics[width=4.27cm]{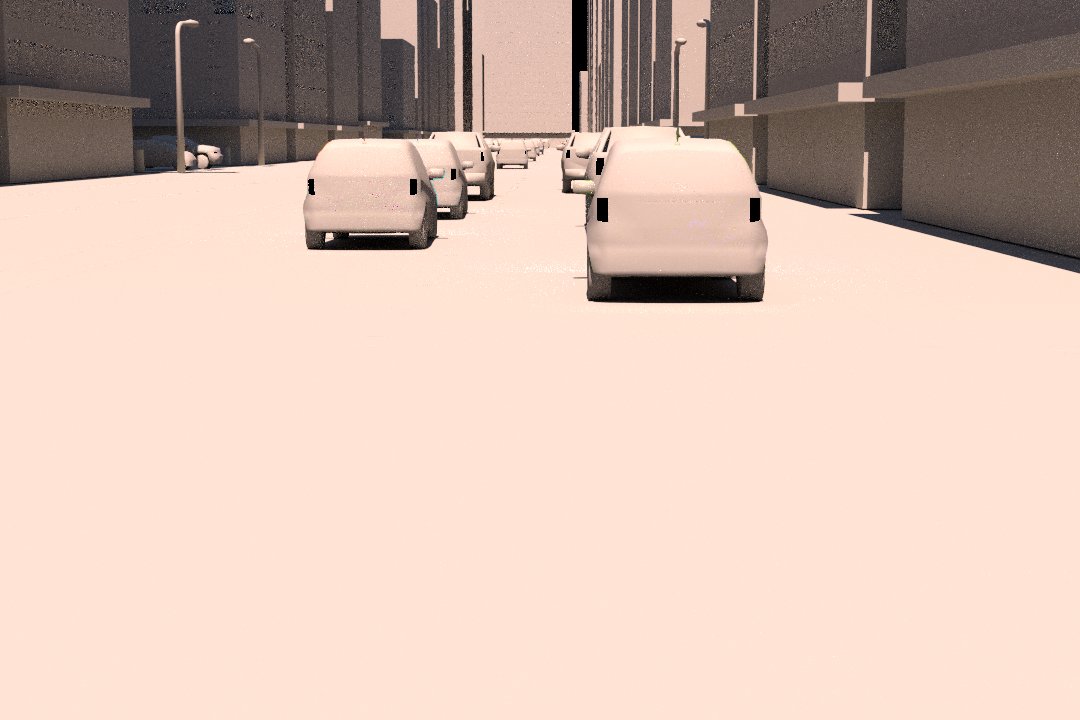}
\includegraphics[width=4.27cm]{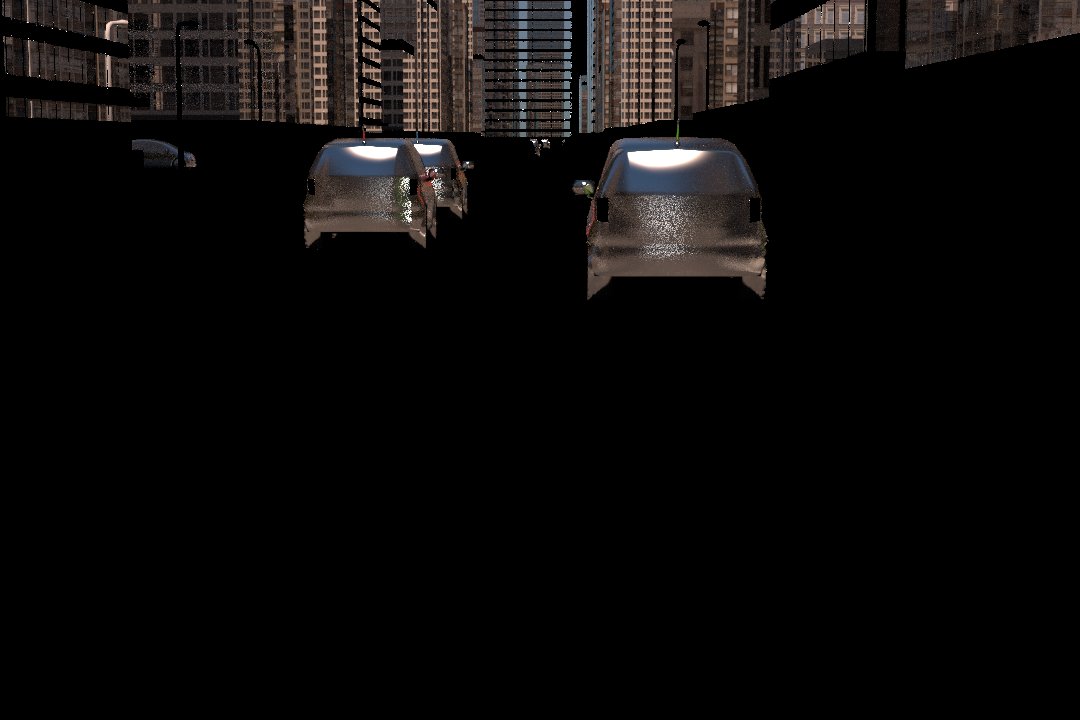}

\includegraphics[width=4.27cm]{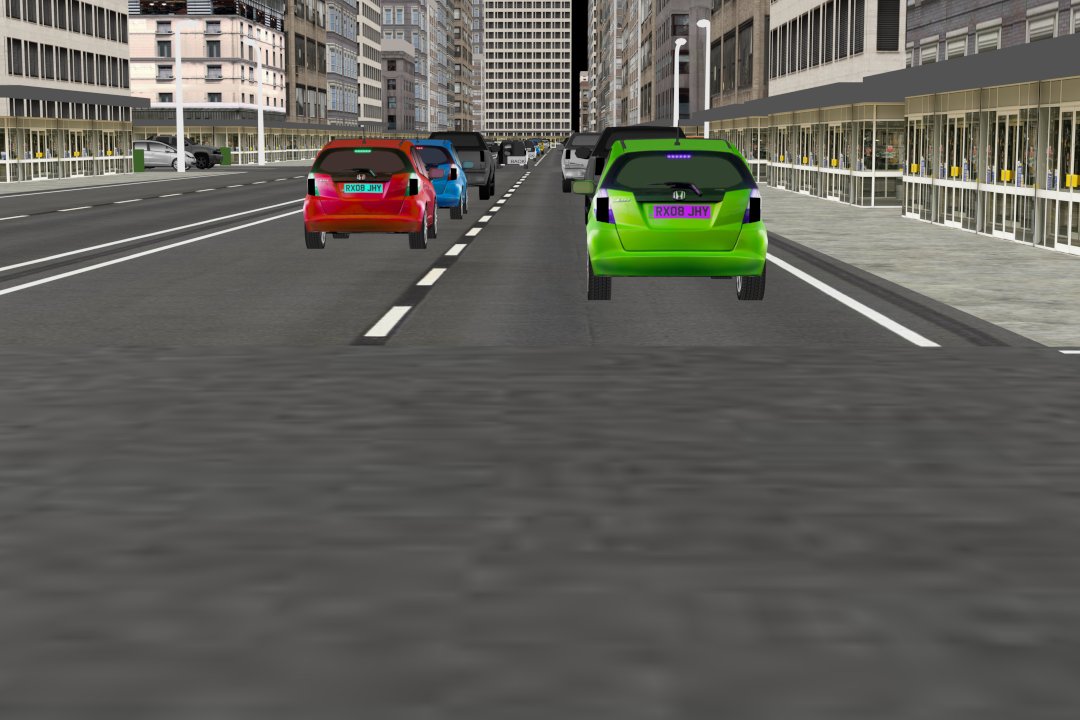}
\includegraphics[width=4.27cm]{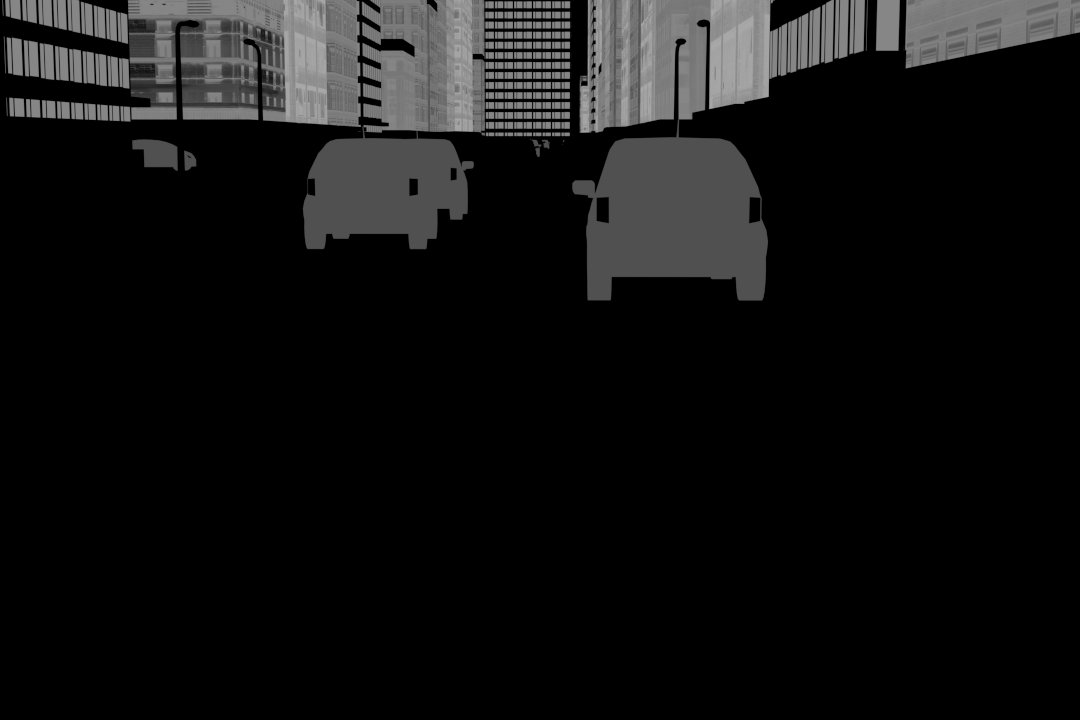}
\includegraphics[width=4.27cm]{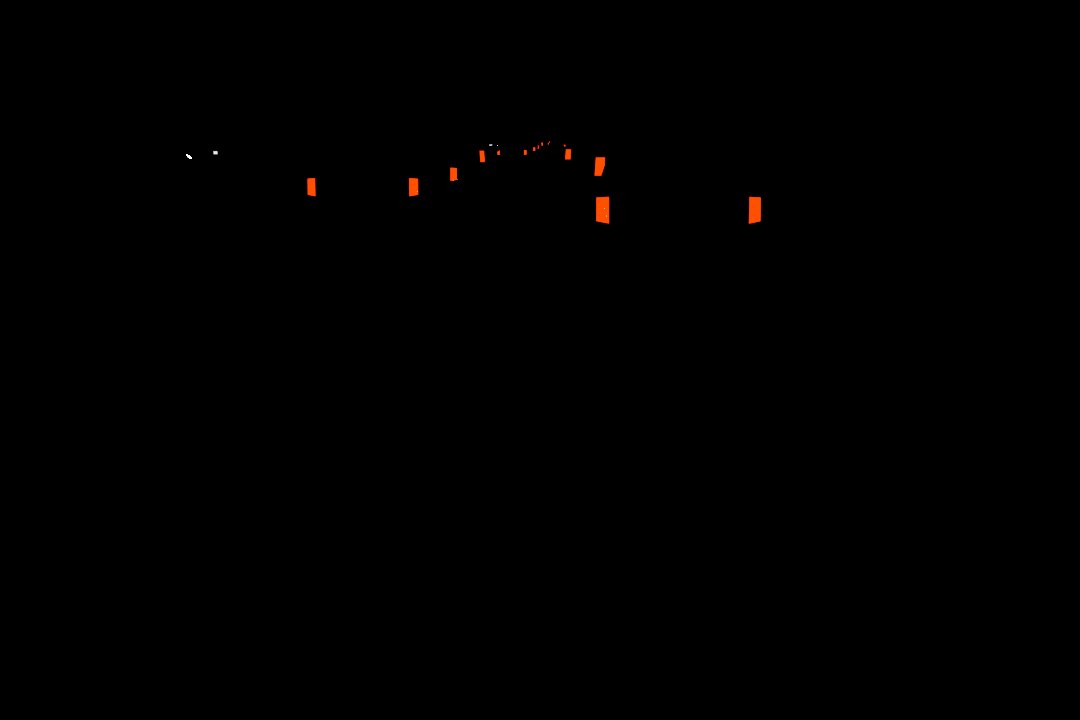}
\includegraphics[width=4.27cm]{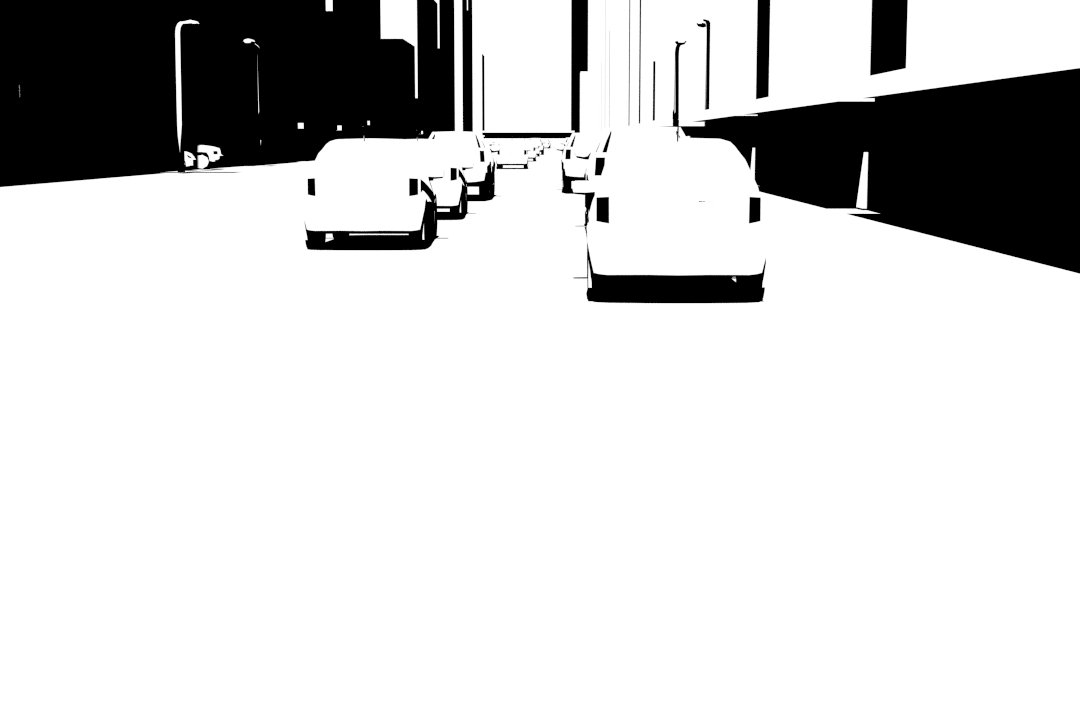}

\includegraphics[width=4.27cm]{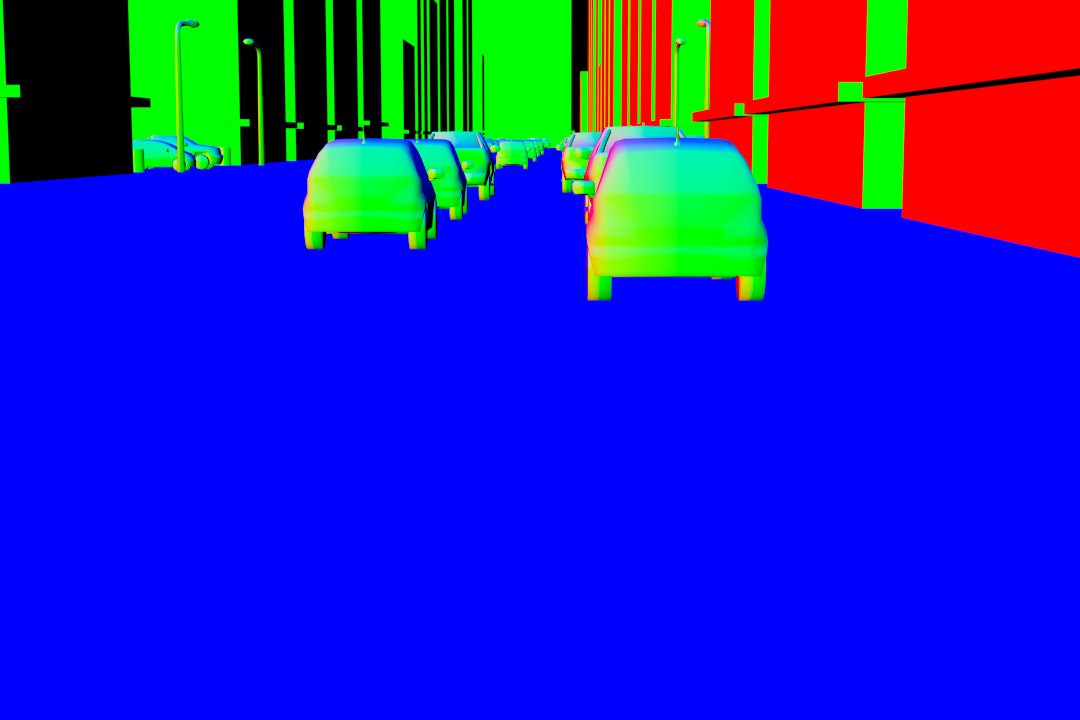}
\includegraphics[width=4.27cm]{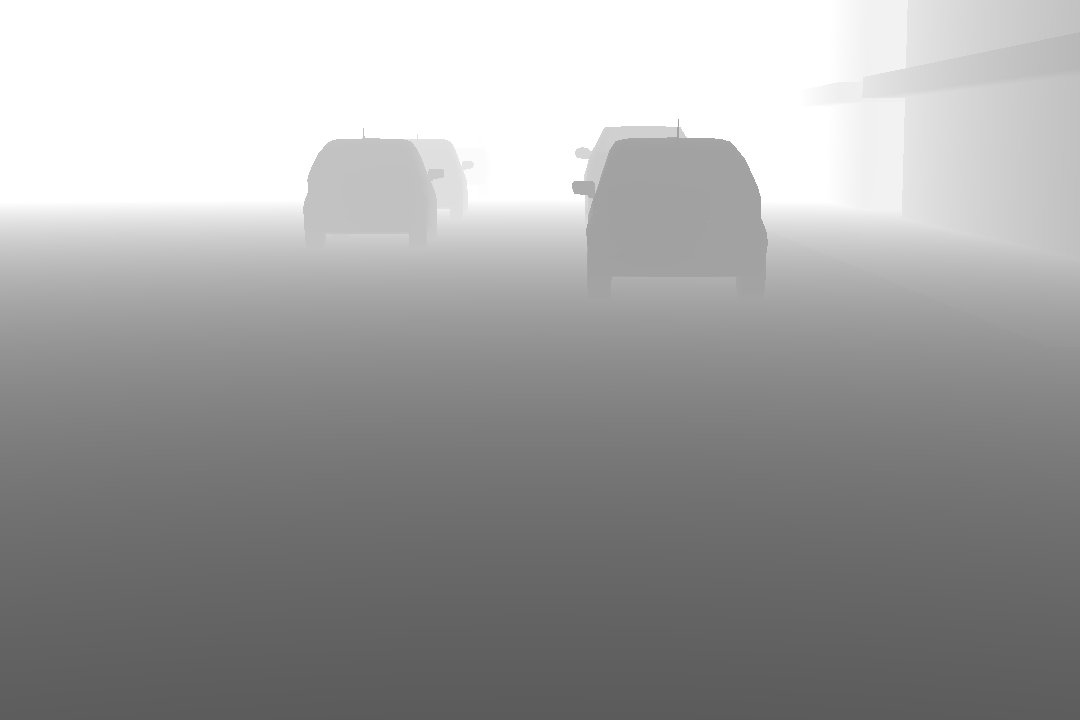}
\includegraphics[width=4.27cm]{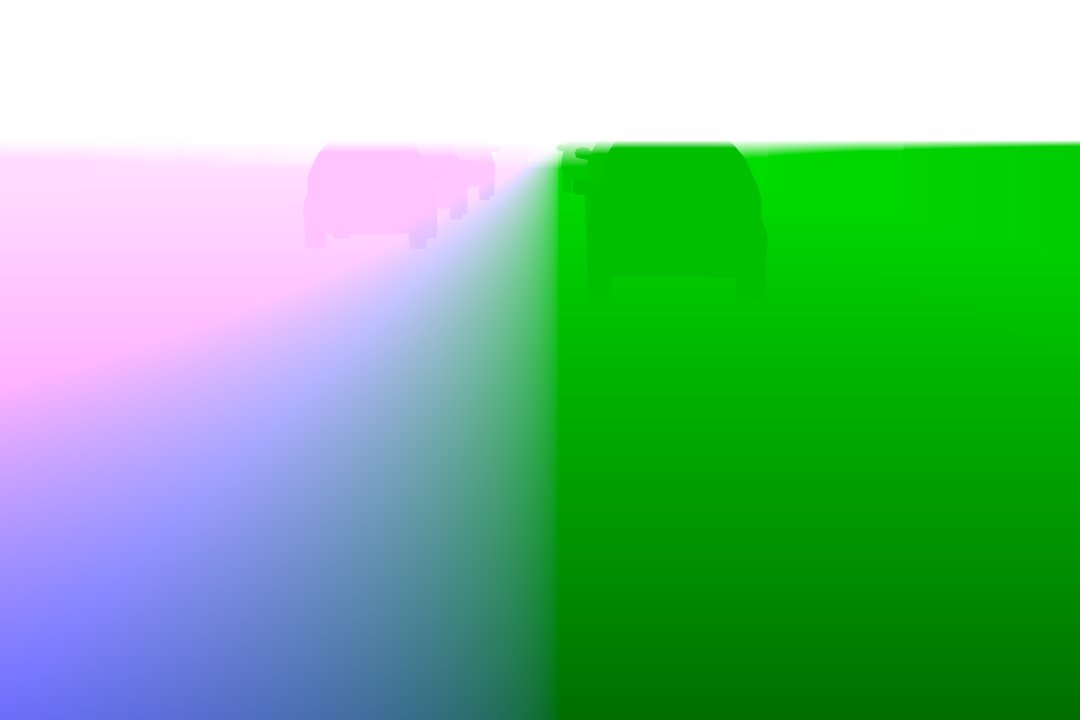}
\includegraphics[width=4.27cm]{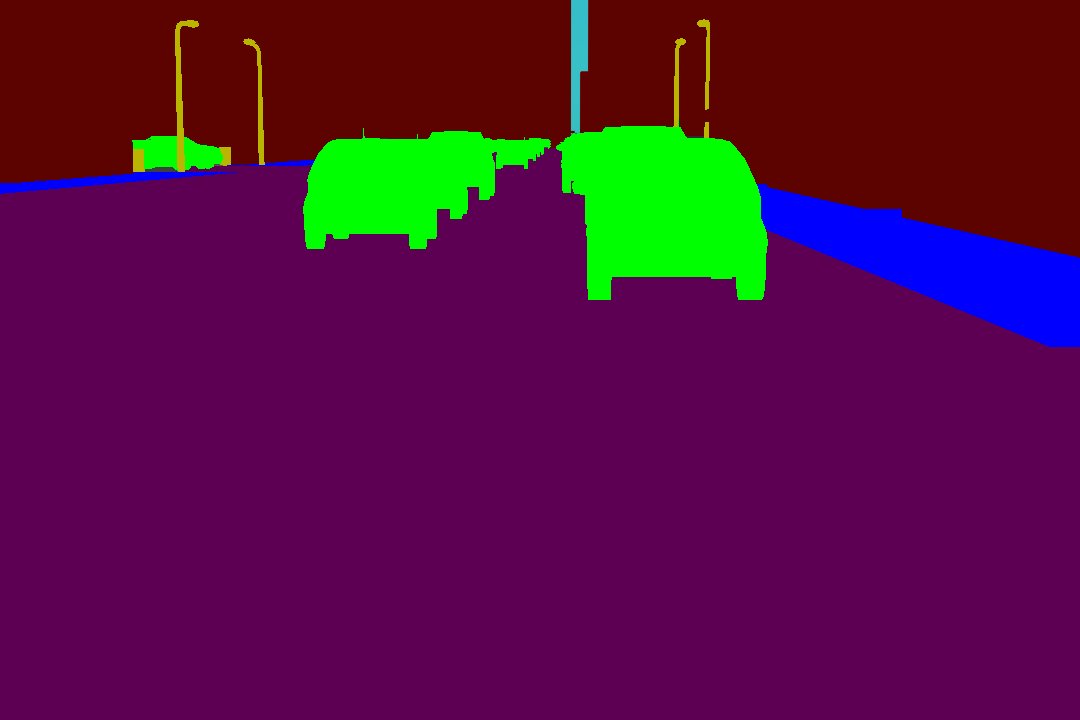}

\caption{\small Annotations and groundtruth: \textit{top row} : RGB, ambient, diffuse, specular shading components; \textit{middle row}: diffuse, specular, emitting surfaces and shadow regions; \textit{bottom row}: surface normals, depth, pixel flow, semantic labels.  }
\label{fig_annotations}

\end{figure*}

\textbf{Annotations}: Nature of required annotations depend on intended goal of vision systems, which varies from Overall System evaluation (e.g. Performance assessment, Offline  learning of algorithm, etc.), to Component evaluations and Model learning (ideal signals and perturbations). Annotations can be  bounding boxes around objects, specifying groups, identifying edges, region labels, etc.  
Annotations in systems engineering context are essentially  used for contextual model learning. Hence, we implement a multitude of procedural shaders which are responsible for rendering multiple image modalities or groundtruth (from local level to semantic level representations) including \textit{depth, surface normals, reflection, shading, diffuse, specular, direct light components, optical flow, geometric flow, trajectories, semantic labels, bounding boxes, shadows} etc. A sample of these modalities are provided in Figure \ref{fig_annotations}.

\textbf{Exposing the parameters}: Any forward rendering method projects 3D scene onto 2D image representations depending on rendering function, which is governed by $I_m = R_m(W_\theta, X_\theta)$, where $I_m$ is the modality or image features, $R_m$ is corresponding rendering kernel program with control parameters $X_\theta$ and $W_\theta$ is set of scene parameters which govern world state including weather, camera and light sources. Our framework exposes all the parameters to the scripting interface (with default values) to make them to be guided by the user or sampled from the probability distributions. The control over both type of parameters enable us to analyze and model the affects of a perturbation (may be weather or illumination variations) on the pixels or to map 3D scene priors to 2D image priors (for instance, conditional distribution of object's size for given camera height from the ground).

\textbf{Limitations of the framework}: In the current framework, object's local coordinated systems are constrained to be coherent with world's coordinate system (i.e object's euler angles are fixed) which means it can only generate Manhattan worlds. As the object models and textures we have are limited, one can observe lot of redundant structures in the generated scenes. 

We can use this platform to analyze the trade-offs and breaking points (performance bounds) of the given estimator by empirically establishing its performance as a function of contextual variables (such as scene geometry, materials, light, weather, and camera etc.) and estimator's tunable parameters (such as scale and thresholds etc.).

\section{Data for RO Model Validation}   \label{sec_data_generation}
\subsection{Simulated data}
\begin{figure*}
\subfloat[Manhattan city scene] {\includegraphics[width=17cm, height=4.7cm]{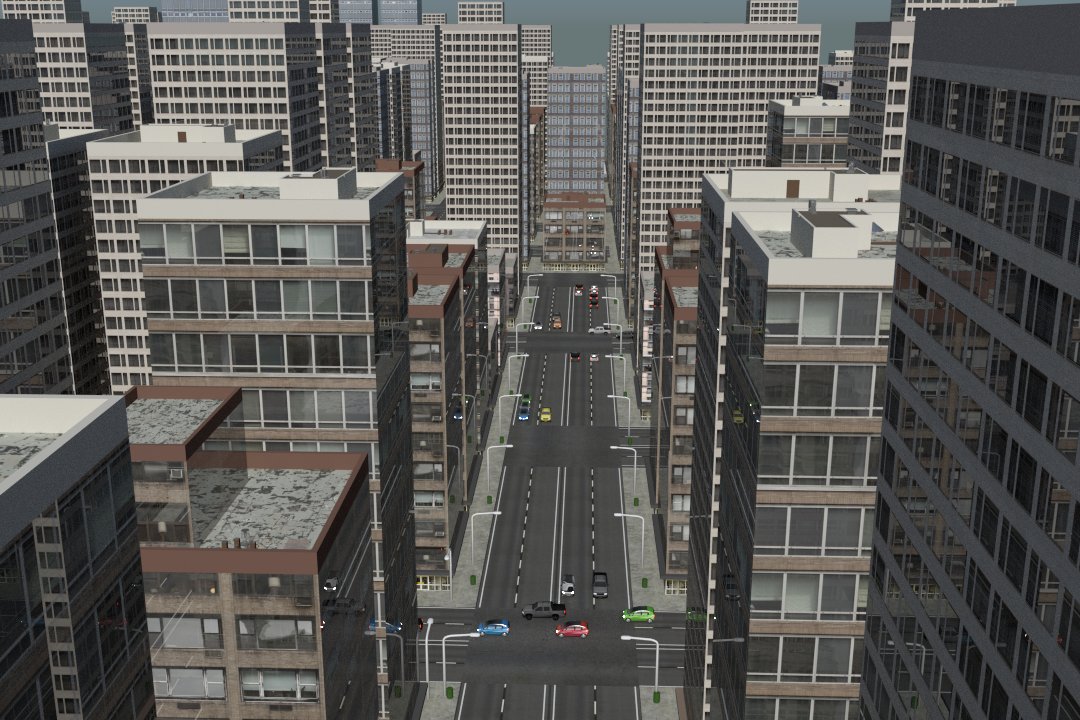}} \\
\subfloat[Global light intensity variations]{
\includegraphics[width=4.2cm]{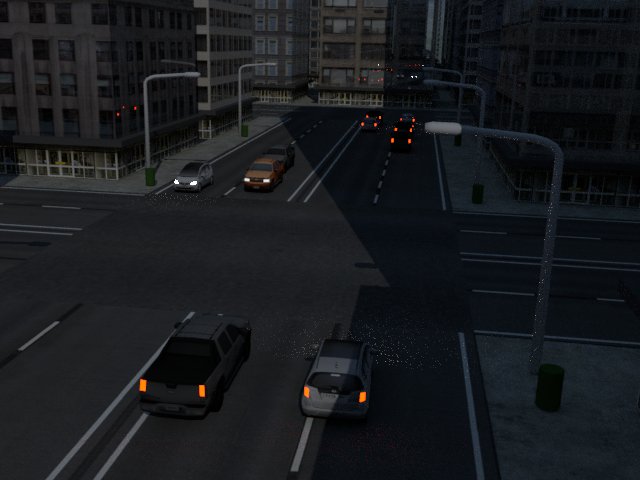}
\includegraphics[width=4.2cm]{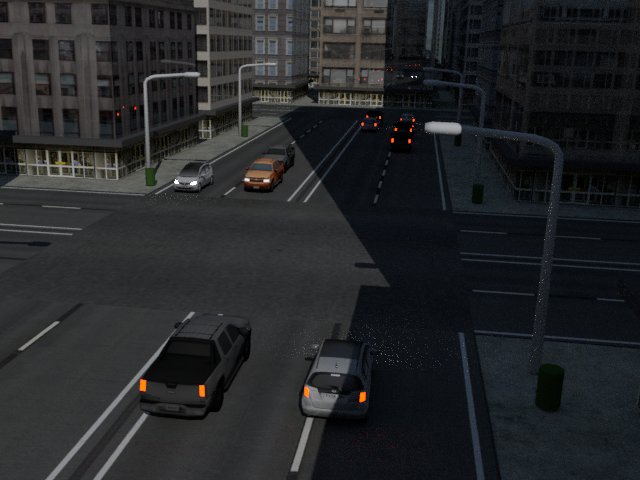}
\includegraphics[width=4.2cm]{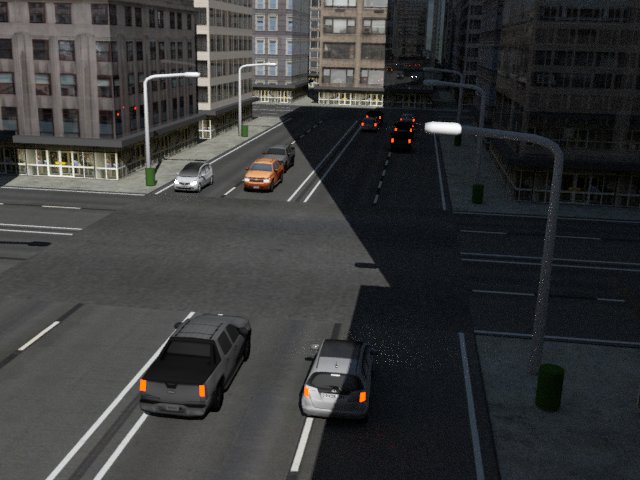}
\includegraphics[width=4.2cm]{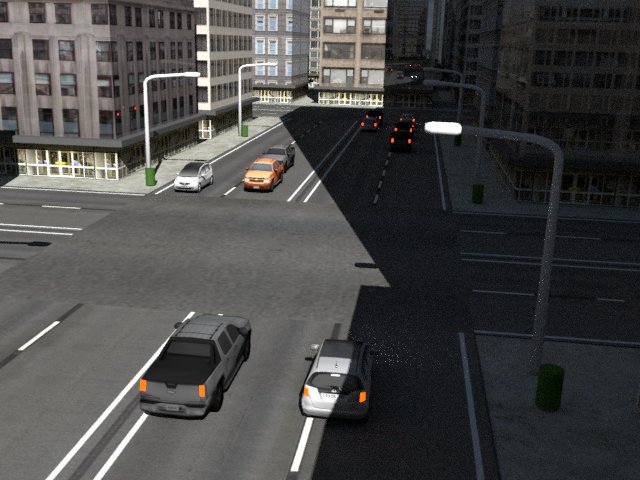}
} \\
\subfloat[Night light variations]{
\includegraphics[width=4.2cm]{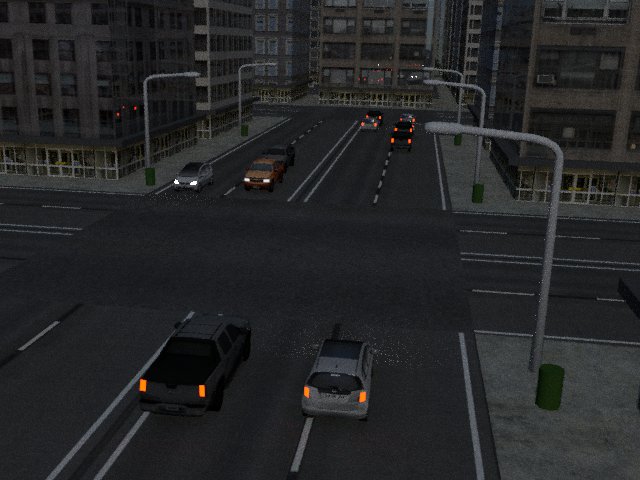}
\includegraphics[width=4.2cm]{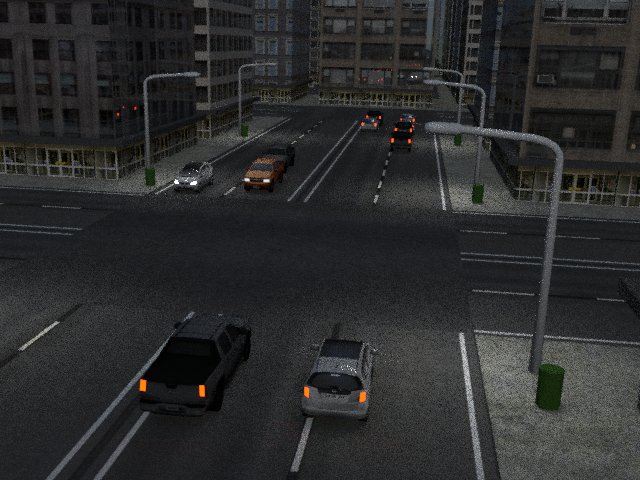}
\includegraphics[width=4.2cm]{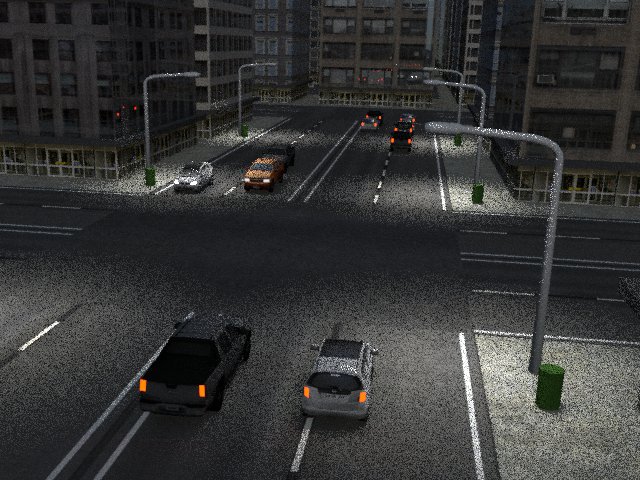}
\includegraphics[width=4.2cm]{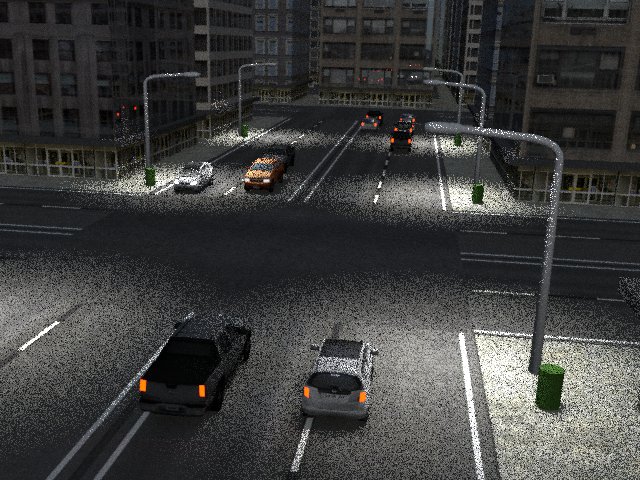}
} \\
\subfloat[Weather variations\label{fig_bad_weather_sim}]{
\includegraphics[width=4.2cm]{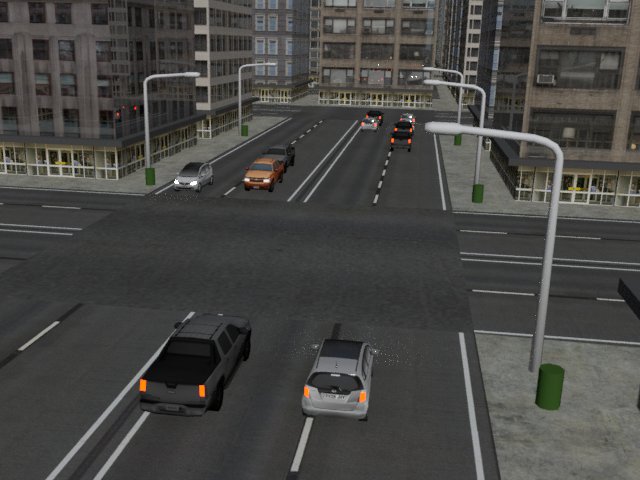}
\includegraphics[width=4.2cm]{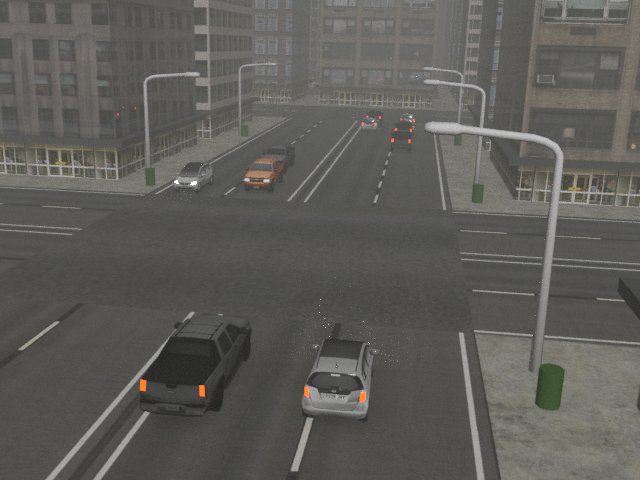}
\includegraphics[width=4.2cm]{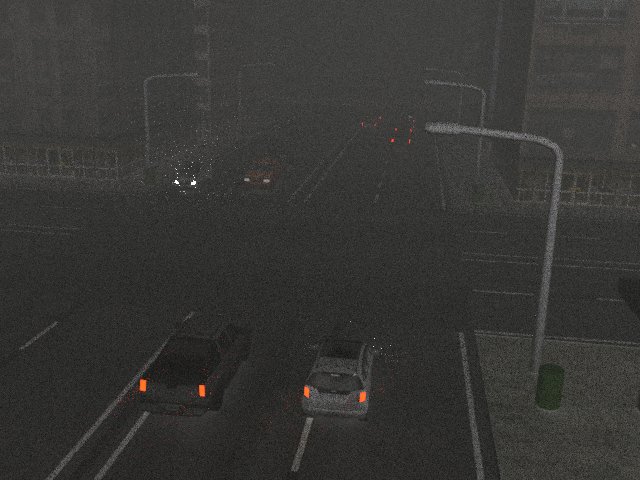}
\includegraphics[width=4.2cm]{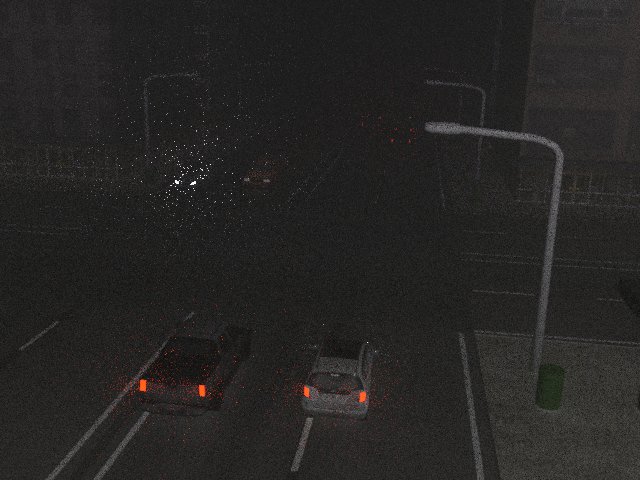}
}
\caption{Simulated samples}
\label{fig_data_simulated}
\end{figure*}

We synthesized a series of images from a Manhattan 3D scene, under different temporal contexts such as global light changes, local light changes (night) and bad weather changes. We used MC path tracing method (with 200 render samples) to render the images used in this work. Some samples are shown in Figure \ref{fig_data_simulated}.

%\subfloat[Validation of RO model under global illumination change \label{fig_model_oc_global}]    {\includegraphics[width=18cm,height=5cm]{images/app_domain/validate_OC_global_light_change_sim.png}} \\

\subsection{Real World data}
\begin{figure*}
\centering
\subfloat[Global light intensity variations, selected from \cite{shimada2014case}]{
\includegraphics[width=4.2cm]{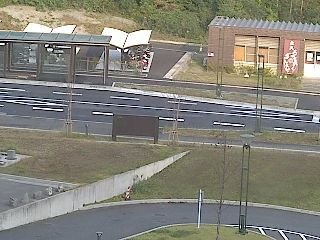}
\includegraphics[width=4.2cm]{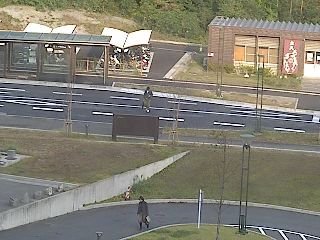}
\includegraphics[width=4.2cm]{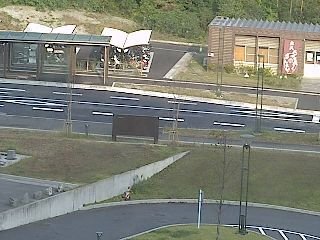}
\includegraphics[width=4.2cm]{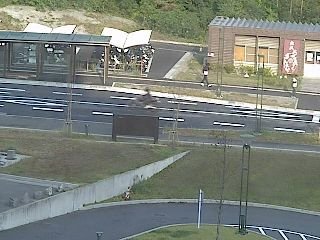}
} \\
\subfloat[Night light variations, selected from \cite{goyette2012changedetection}]{
\includegraphics[width=4.2cm]{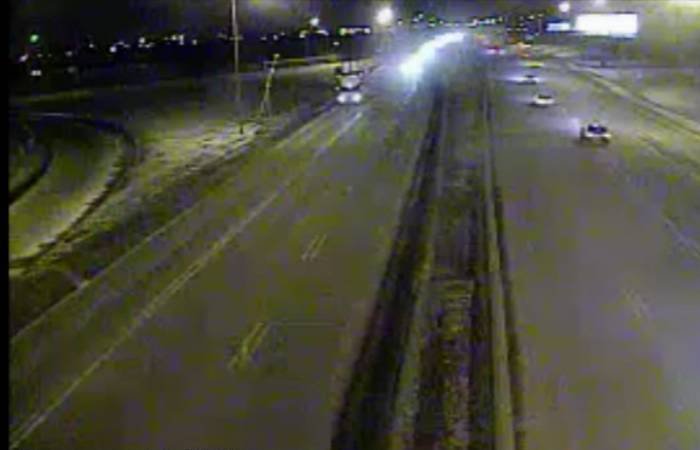}
\includegraphics[width=4.2cm]{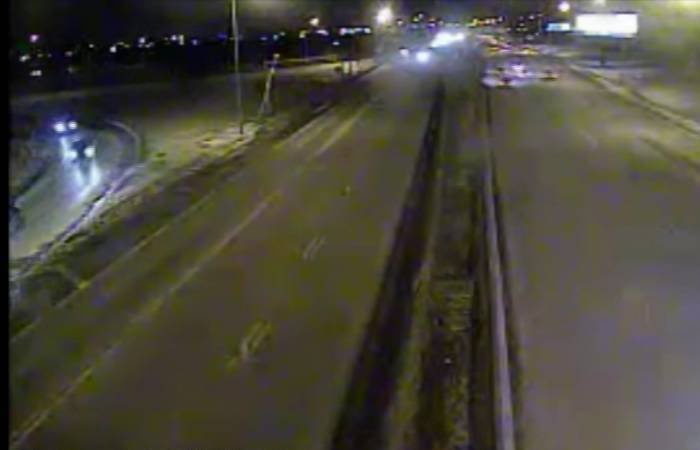}
\includegraphics[width=4.2cm]{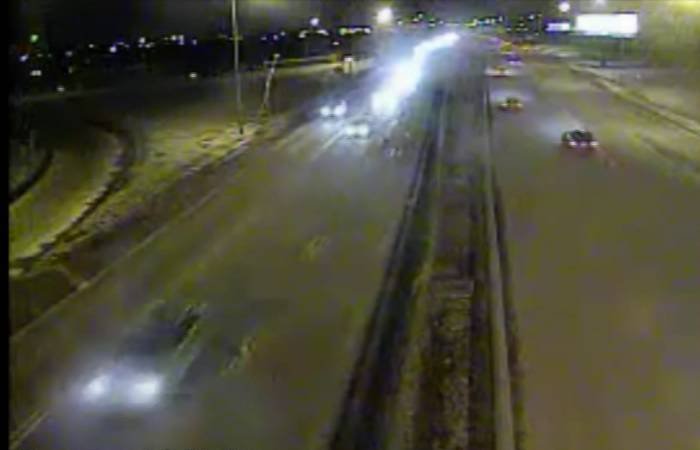}
\includegraphics[width=4.2cm]{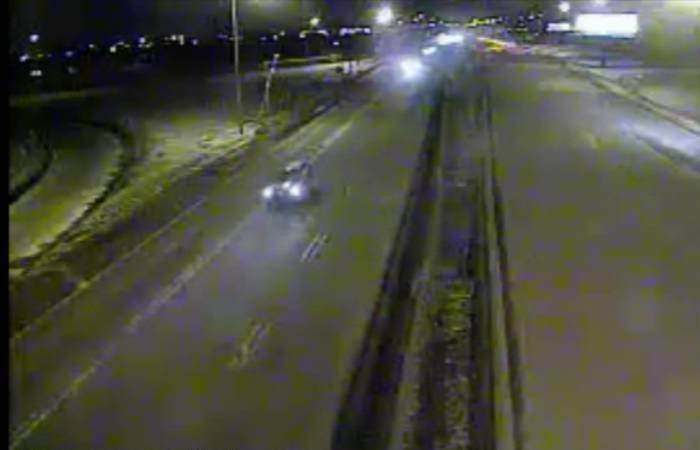}
}\\
\subfloat[Bad Weather variations, selected from \cite{goyette2012changedetection}]{
\includegraphics[width=4.2cm]{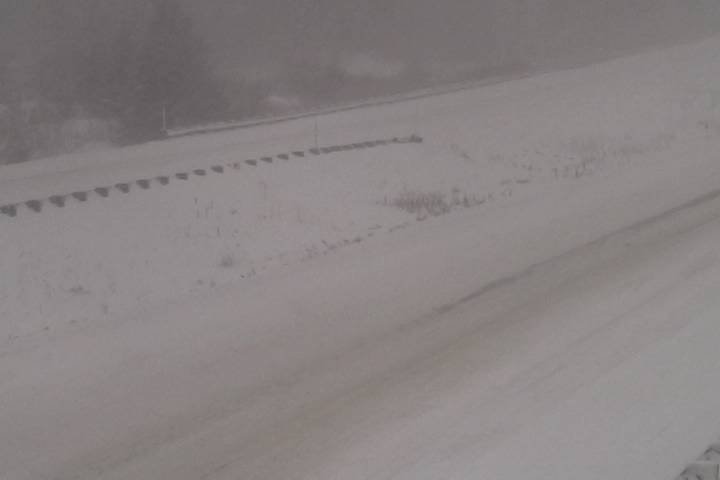}
\includegraphics[width=4.2cm]{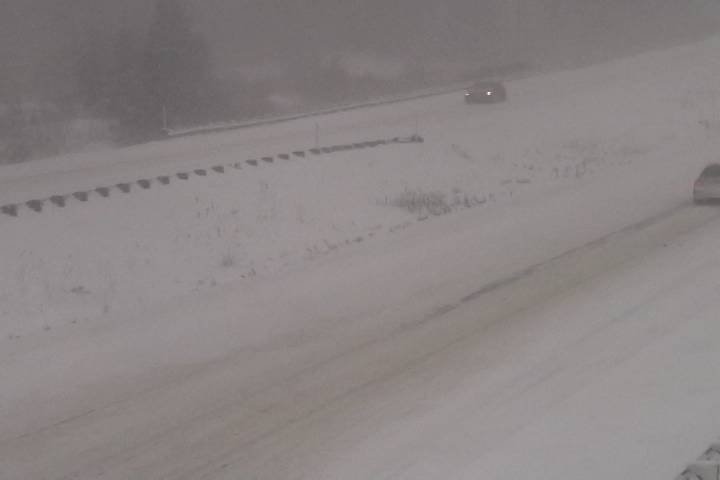}
\includegraphics[width=4.2cm]{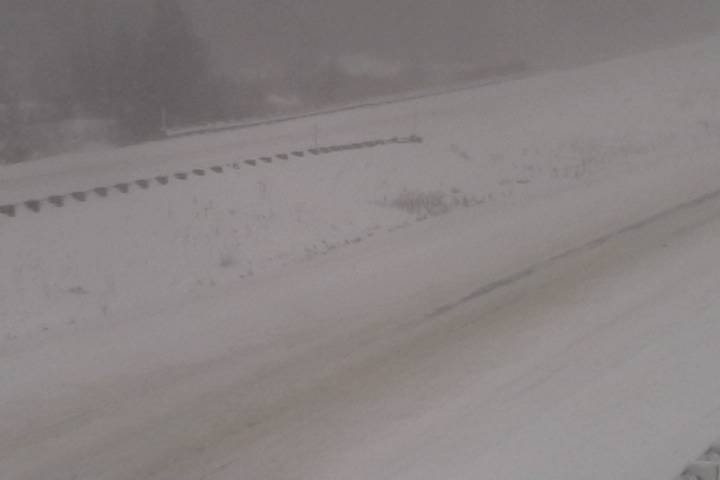}
\includegraphics[width=4.2cm]{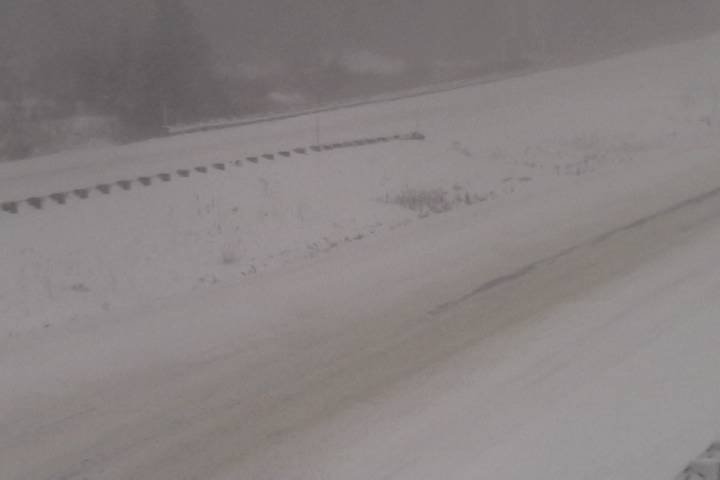}
}
\caption{Real world data samples}
\label{fig_data_real}
\end{figure*}

We have chosen some real world video sequences from change detection benchmark datasets \cite{shimada2014case,goyette2012changedetection}, which are captured under different temporal variations, similar to the ones considered above. Samples are shown in Figure \ref{fig_data_real}. 

\section{Change Detection Experiments} \label{sec_change_detection}
In this section, we provide motivations for the use of quasi-invariant transforms validated in the experiments, discussed in the main article. We also include additional experiments and discussions. 

{\bf Sub-Space Analysis-DCT: } One prominent way to measure the robustness of input signal is the compactness of the data in the projected sub-space. The key challenge is making the choice of projection operations with balance in the tradeoff of minimal signal loss and target invariance \cite{bruna2013invariant}. For background modeling and effective representation the intra-class spread of the background has to be small compared to foreground distributions. In data compression domain, DCT (Discrete Cosine Transform) is very effective in preserving essential frequency components for reconstruction of the data and further more it is optional to choose the number of coefficients as per the required reconstruction quality hence appropriate number of frequency components. Further, It is reasonable to assume that various foreground objects would spread widely in the sub-space. This implies that, with a proper selection of frequency coefficients, it would be possible to approximately describe the background signal by using only a lesser number of coefficients. One of the important question is selection of representation for DCT transform, 2D image patch, or overlapping patches or 3D video bricks, which depends upon scenarios. In order to build background model on videos, we studied the robustness properties of the  DCT response by empirical evaluation in Section \ref{Validation}.

{\bf Rank based Ordinal measure:}  Rank order is proposed as a robust correlation measure using only the ordering of the pixel values rather than the absolute intensity. Use of ordinal measures between two image patches is a powerful approach and is defined as the distance metric between their rank permutations \cite{bhat1996ordinal}. As it is not dependent on the gray values, it is insensitive to data outliers and invariant to global illumination change or camera gain.  Experimental results also proven the performance of the approach over other correlation based methods like SSD (Sum Squared Differences) and NCC (Normalized Correlation Coefficients). SSD has computational advantages over NCC, where as NCC is preferred because of its invariance property to linear brightness and contrast variations. Image transform methods such as rank transform methods  \cite{bhat1998ordinal, heikkila2004texture}, are also applied for image matching where the correlation of transformed image is not dependent upon pixel values so these methods are relatively insensitive to presence of data outliers,  and monotonic transformations like gamma correction. The draw back of this method are it is dependent on the reference pixel (e.g. center pixel), with implicit assumption of surface being smooth and lambertian, and not robust to window distortion like projective distortion and cyclic shift. 

\begin{figure*}[!hbt]
\centering
\subfloat[Gaussian]{ \includegraphics[scale=0.3]{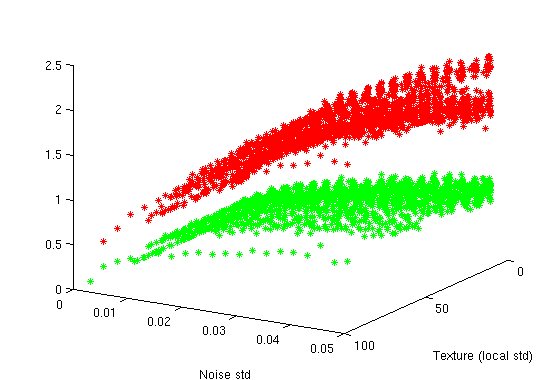} }
\subfloat[Salt and Pepper]{ \includegraphics[scale=0.3]{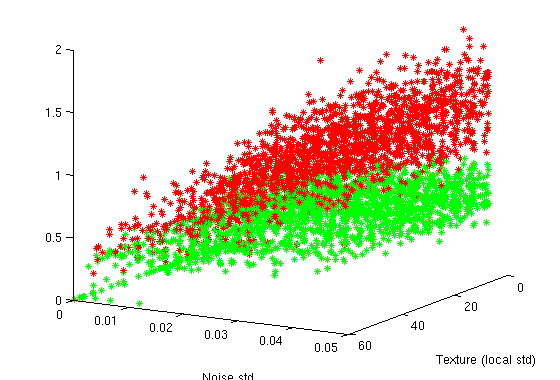}}
\subfloat[Speckle]{\includegraphics[scale=0.3]{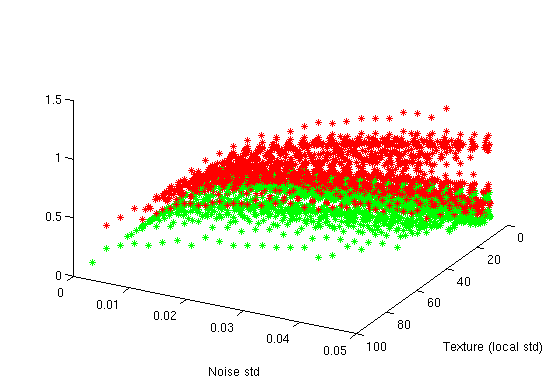}} \\
\subfloat[Gaussian]{ \includegraphics[scale=0.3]{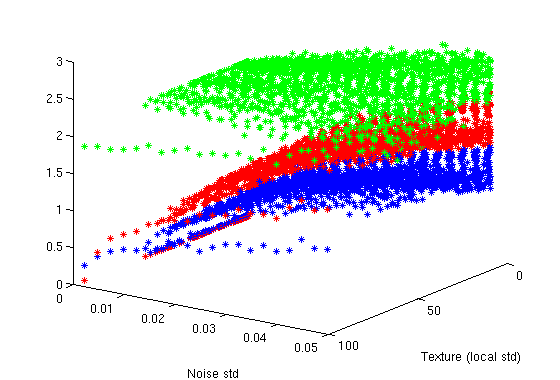} }
\subfloat[Salt and Pepper]{ \includegraphics[scale=0.3]{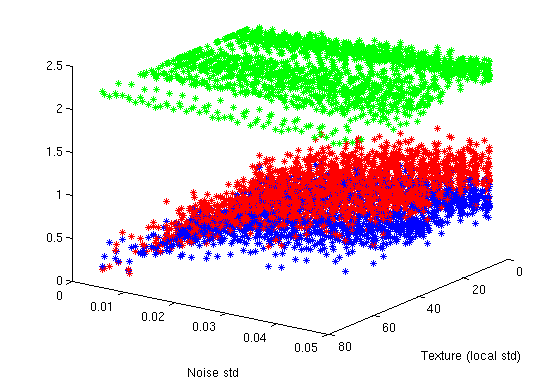}}
\subfloat[Speckle]{\includegraphics[scale=0.3]{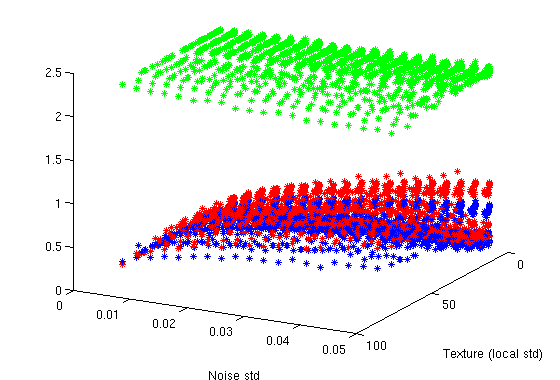}}
\caption{DCT and RO robustness w.r.t Noise and  Texture - Simulated Data: RO response is presented in red color, while green represents DCT measures and their combination (DCT+RO) is shown in blue.}%, Illumination and blur }
 \label{DCTRORobustness_simulation}
 \end{figure*}
 
 \begin{figure*}
\centering
\subfloat[RO based change detection]{\includegraphics[width=17cm,height=3.7cm]{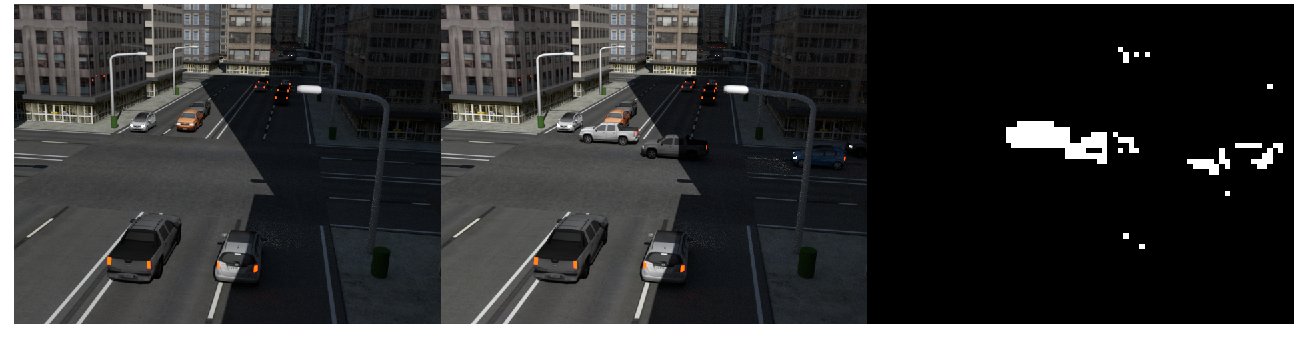}}\\
\subfloat[DCT based change detection]{\includegraphics[width=17cm,height=3.7cm]{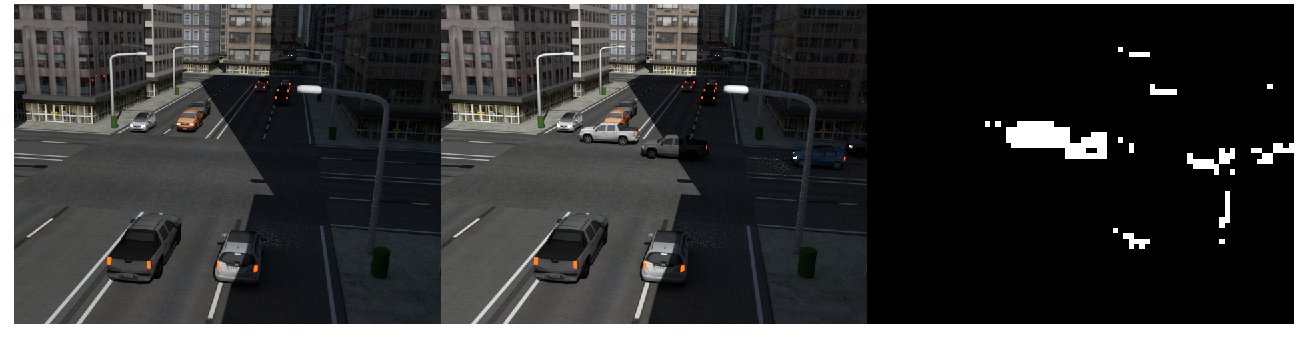}}\\
\subfloat[DCT+RO based change detection]{\includegraphics[width=17cm,height=3.7cm]{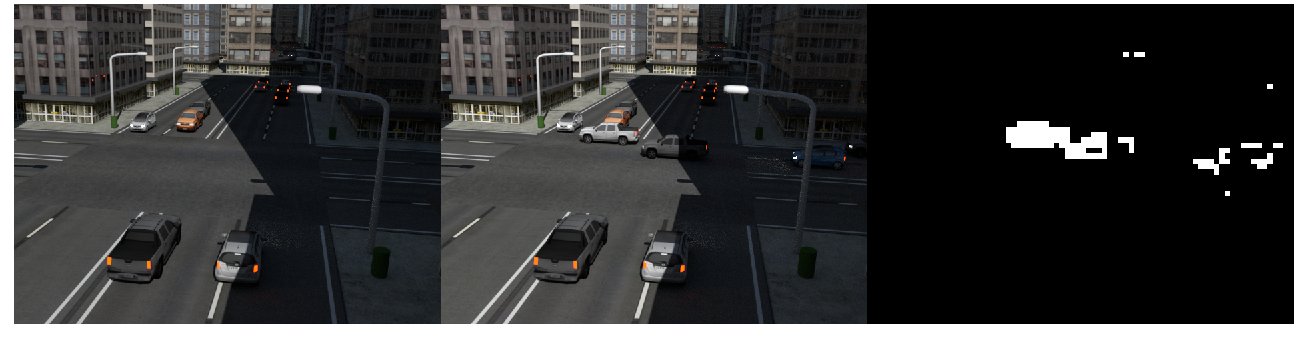}}
\caption{Change detection results}
\label{fig_changedetection}
\end{figure*}

\subsection{Performance modeling of Invariant Modules} \label{Validation}
{\bf DCT - Subspace:}
 As explained in previous section we first apply DCT on the background image and input image to extract the block coefficients. These block coefficients corresponding to radial frequency energies for the considered images.  The radial frequency energies are scaled by a ratio of a first block coefficient to the other image block coefficients. Then we compute the energy difference for at least one of the radial frequency energies between the images. The number of frequencies to be used in the difference computation is a parameter for the algorithm and its effects we left for future study.  The  detection of scene change flagged if the energy difference is above the indicative threshold computed from background test statistics. 
  
%The effectiveness of the DCT analysis on noise and illumination perturbations is shown in Figure \ref{DCTRORobustness_simulation} to depict how the methods works for illumination invariant change detection directly in the discrete cosine transform (DCT) space and Rank order matching on these DCT coefficients. The illumination change is assumed as a local contrast change, and treated with a nonparametric rank matching of the DCT coefficients \cite{zoghlami2005illumination}.  
  
{\bf Ordinal measure:}
To match two images with rank order, Lets consider $\pi_1$, $\pi_2$ are rankings of image $I_1$ and $I_2$ respectively with size of n pixels and $d_h$ is the hamming distance, and $S_n$ be the set of all permutation of integers $[1, 2, 3, ... ,n ] $. The distance metric between two permutations is $d(\pi_i,\pi_j) $ for $\pi_i, \pi_j$ $\in$ $S_n$. One such distance metric is Hamming distance $d_h (\pi_1,\pi_2) = \sum_i (\mid sign(\pi_1^i,\pi_2^i)\mid)$. Similar methods are used in literature, where matching between two descriptors is based on a distance function that penalizes order flips. 

Mittal et.al \cite{mittal2006intensity} extended the work on ordinal measure by incorporating pixel values which penalizes the order flips by their intensities and applied to change detection. Because of the inclusion of intensity value and by computing statistics over them, it is more robust to intensity noise. In the work by Maneesh Singh et. al. \cite{singh2008order}, they have shown superior results compared to prior methods by incorporating noise statistics. In their work they applied statistical test for order consistency with presence of noise with know statistical property where they compute L2 distance (similar to the approach proposed in Mittal \& Ramesh \cite{mittal2006intensity}) by projecting the second patch on the rank-set boundary of the first patch. The distance between blocks from background image ($I_b$)  and current image ($I_c$) is computed as $
  \hat {d}_{RO}= \frac{1}{N} {\parallel(Q_b - \hat Q_c)\parallel}^2. 
$
 where,  $Q_b$ is background block and $ \hat Q_c$ is rank consistency block estimated from image $I_c$ of size $N$. It is proven to be robust against Gaussian noise. 
 
 The effectiveness of the modules (DCT, RO, and their combination) and their responses with added noise and illumination perturbations is as shown in Figure \ref{DCTRORobustness_simulation}. It depicts that how these methods work for illumination invariant change detection. The combination of DCT and RO (DCT+RO) is done by applying rank order matching on computed DCT coefficients. The illumination change is assumed as a local contrast change, and treated with a nonparametric rank matching of the DCT coefficients \cite{zoghlami2005illumination}.  The response of the combination  is lesser compared to that of individual modules (blue plots in Figure \ref{DCTRORobustness_simulation}), which proves its robustness towards the considered perturbations.

\textbf{Change detection results}: We applied the combination (DCT + RO) of operations for change detection in illumination change data and results are shown in Figure \ref{fig_changedetection}. In the first row we have shown change detection results form RO, DCT result is shown in second row and in the third row the detection of of the combined operation (DCT+RO) is shown. The effect of noise and illumination are quite visible in the obtained results while applying DCT and RO separately, while the combination operation produces better result in the considered case.

\section{Conclusion}
We provided the details of the graphics simulation platform, motivated by the need for validation or characterization of the vision models and modules in the context.  
We have demonstrated the combination of classical simulation
based testing in performance characterization literature
in combination with graphics simulated data. The main
insight in this part of work reinforces the points that simulation
based testing gives different types of insights, i.e. qualitative as well
as quantitative in nature. Another point that is expressed
above is, the view that such characterizations mainly allow
establishment of correspondence between model spaces
(e.g. physically motivated models and generative models)
and unraveling these correspondences provide better understanding
of limits of algorithms and in addition may spur
interesting explorations in modeling in the related fields.

\end{document}